\documentclass[10pt,twocolumn,letterpaper]{article}

\usepackage{cvpr}   

\usepackage{multirow}
\usepackage{booktabs} 
\usepackage{array}    
\usepackage{adjustbox} 
\usepackage{cuted}
\usepackage{xcolor}
\usepackage{tikz}
\usepackage{fontawesome}
\usepackage{amsmath}
\usepackage{bm}
\usepackage{caption}
\usepackage{placeins} 
\usepackage{dblfloatfix} 
\usepackage{subcaption} 
\usepackage{mathtools}
\usepackage[table]{xcolor}
\usepackage{enumitem}
\usepackage{newunicodechar}
\newunicodechar{≤}{\ensuremath{\leq}}
\usepackage{fancyvrb}
\usepackage{fvextra}
\usepackage{xcolor}
\usepackage{listings}

\definecolor{codebg}{RGB}{248,248,248}
\definecolor{codecomment}{RGB}{0,128,0}
\definecolor{codekeyword}{RGB}{0,0,180}
\definecolor{codestring}{RGB}{163,21,21}

\definecolor{userblue}{HTML}{0088EE}
\newcommand{\yellow}[1]{\textcolor{yellow}{\textbf{#1}}}
\newcommand{\blue}[1]{\textcolor{userblue}{\textbf{#1}}}
\newcommand{\red}[1]{\textcolor{red}{\textbf{#1}}}
\newcommand{\gray}[1]{\textcolor[gray]{0.2}{\textbf{#1}}}

\definecolor{cvprblue}{rgb}{0.21,0.49,0.74}
\usepackage[pagebackref,breaklinks,colorlinks,allcolors=cvprblue]{hyperref}

\def\paperID{4380} 
\def\confName{CVPR}
\def\confYear{2026}

\title{Anchoring and Rescaling Attention for Semantically Coherent Inbetweening}

\author{
Tae Eun Choi\thanks{Equal contribution.}\hspace{1.5em}
Sumin Shim\footnotemark[1]\hspace{1.5em}
Junhyeok Kim\hspace{1.5em}
Seong Jae Hwang\\
Yonsei University\\
{\tt\small \{teunchoi,use08174,timespt,seongjae\}@yonsei.ac.kr}
}

\begin{document}
\maketitle
\begin{strip}
  \centering
  \includegraphics[width=\linewidth]{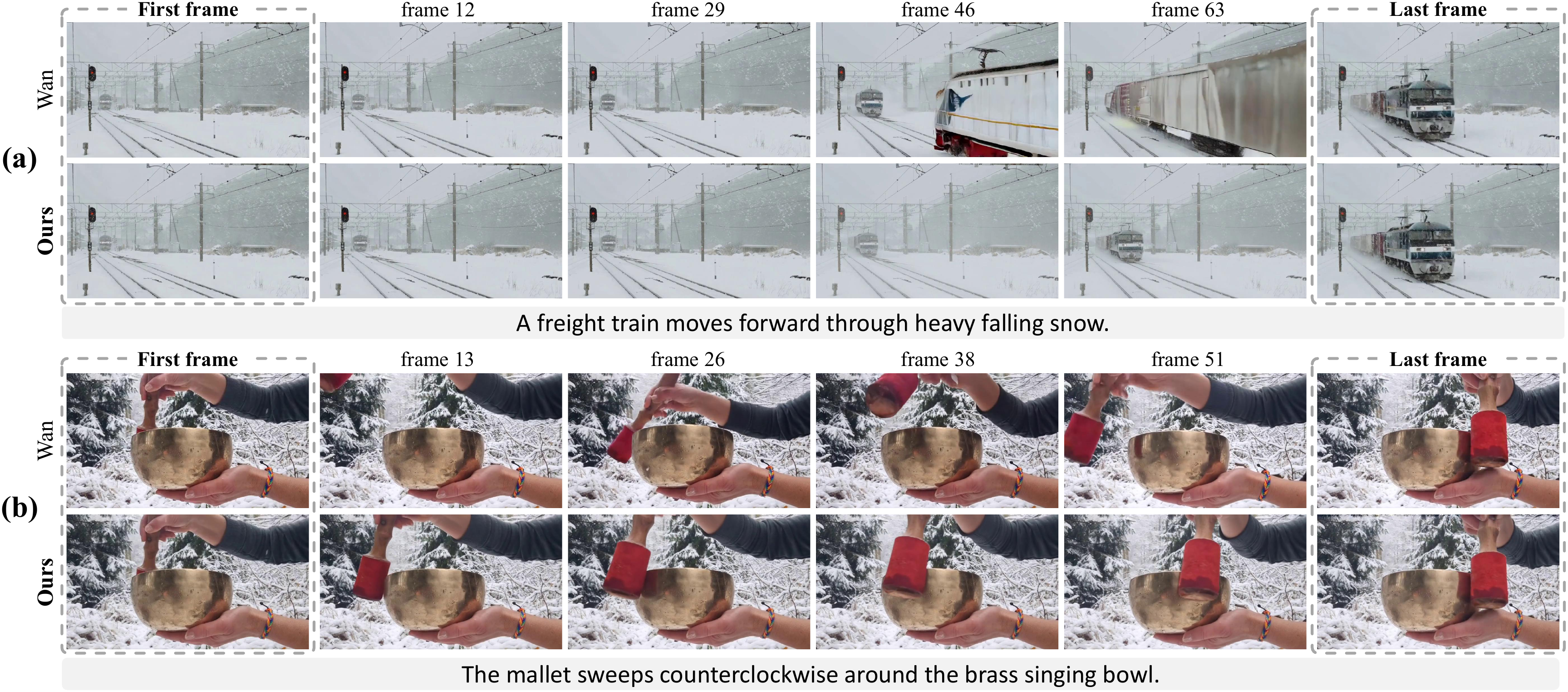}
  \captionof{figure}{
  We introduce a training-free approach on the task of \textit{generative inbetweening} which generates intermediate frames using the two keyframes and text. In \textbf{(a)}, our method correctly recognizes the train and produces consistent and coherent frames. In \textbf{(b)}, we improve semantic alignment between the text and generated frames, accurately capturing the `counterclockwise' movement, in contrast to Wan~\cite{wan}.
}
  \label{fig:teaser}
\end{strip}

\begin{abstract}
\vskip -3mm

Generative inbetweening (GI) seeks to synthesize realistic intermediate frames between the first and last keyframes beyond mere interpolation.
As sequences become sparser and motions larger, previous GI models struggle with inconsistent frames with unstable pacing and semantic misalignment.
Since GI involves fixed endpoints and numerous plausible paths, this task requires additional guidance gained from the keyframes and text to specify the intended path.
Thus, we give semantic and temporal guidance from the keyframes and text onto each intermediate frame through Keyframe-anchored Attention Bias. 
We also better enforce frame consistency with Rescaled Temporal RoPE, which allows self-attention to attend to keyframes more faithfully.
TGI-Bench, the first benchmark specifically designed for text-conditioned GI evaluation, enables challenge-targeted evaluation to analyze GI models. 
Without additional training, our method achieves state-of-the-art frame consistency, semantic fidelity, and pace stability for both short and long sequences across diverse challenges.
\vskip -3mm
\end{abstract}
\vspace{-8pt}
\section{Introduction}
\label{section:intro}
Video frame interpolation (VFI) aims to predict one or more intermediate frames between two keyframes, the first and last input images, often to raise frame rate or enable slow motion \cite{reda2022film, ding2024video, cheng2021multiple}.
Recent advancements in image-to-video models~\cite{svd,wan} enhanced overall video generation quality and enabled longer sequence generation.
Building on this, current works have shifted from matching a single ground truth to generating varied scenes between sparser keyframes.
This task, namely generative inbetweening (GI), reframes VFI as filling the gap between widely-spaced keyframes with plausible, temporally coherent transitions under uncertainty \cite{ace}.

Early works on GI was driven by Stable Video Diffusion (SVD)~\cite{svd,trf,vibid,gi,fcvg}. 
While GI task requires to be conditioned on two keyframes, SVD can structurally take only a single keyframe as an input, inevitably leading prior works to run SVD twice and fuse the results to approximate the intermediate frames.
However, this approach leads to collapse and blur especially on long sequences as GI inherently requires exploiting both keyframes.
In contrast, Diffusion Transformer (DiT)–based video models~\cite{cogvideox, kong2024hunyuanvideo, wan} are able to jointly condition on two keyframes as well as text prompts while scaling to long sequences.
Consequently, recent studies including Wan's First-Last-Frame-to-Video (FLF2V) pipeline began leveraging DiT for text-conditioned GI \cite{wan}.

\begin{figure}
  \centering
    \includegraphics[width=\linewidth]{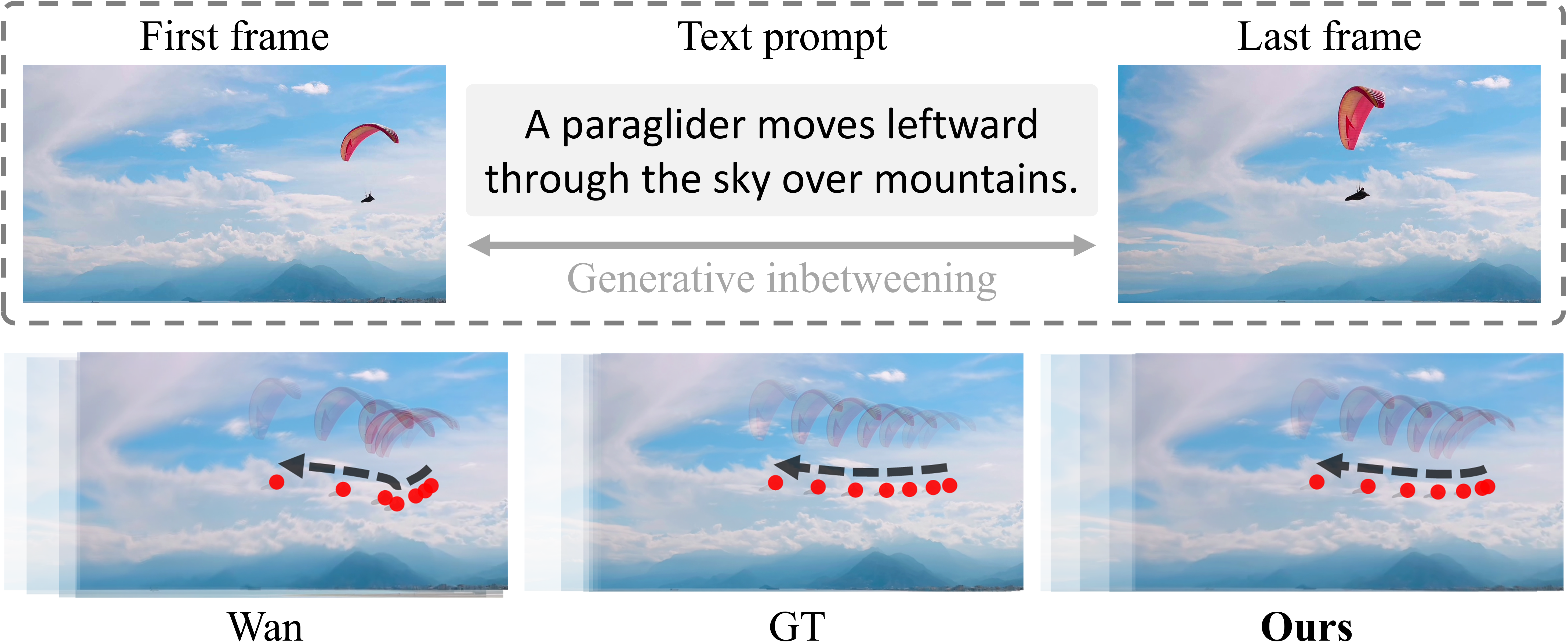}
    \caption{\textbf{Pace Stability Comparison.} This figure compares the pace stability of Wan and our method against ground truth (GT). The paraglider’s motion is visualized by overlaying same, uniformly sampled indices for GT, Wan, and Ours with the background aligned, marking sampled positions with red dots (\red{$\bullet$}) and displacements with black arrows (\gray{$\pmb{\dashleftarrow}$}). Wan exhibits pace instability, in that the paraglider alternately accelerates and decelerates, producing uneven spacing whereas our method closely matches the ground truth with smooth motion and stable pacing.
    }
    \vspace{-8pt}
    \label{fig:temporal}
\end{figure}

As keyframes become sparser and motions more dynamic, the guidance from the two keyframes and text prompt on intermediate frames naturally weakens along the generation process.
We identify these issues as three key challenges: (i) semantic fidelity, (ii) frame consistency, and (iii) pace stability.
For instance, in ~\cref{fig:teaser}, Wan shows text misalignment with inconsistent frames, while ~\cref{fig:temporal} demonstrates an example of pace instability with spontaneous pace shifts.
Consequently, a dedicated mechanism is needed to draw semantic and temporal cues from the two keyframes and text conditions.
We achieve this by modifying the DiT’s cross- and self-attention individually, which are crucial for context mixing~\cite{transformer_circuit,attention} without additional training.

First, we implement Keyframe-anchored Attention Bias (KAB) on cross-attention in order to maintain semantic fidelity and stable pacing.
More specifically, we gain \textit{keyframe anchors} from the keyframes' cross-attention which are used to attract the intermediate frames towards the conditions: the two input keyframes and text prompt.
This guides the intermediate attention maps to fill in the missing semantic and temporal cues from all conditions.
Through this method, KAB yields videos that are semantically aligned and stably paced.

Furthermore, we propose Rescaling Temporal RoPE (ReTRo), a simple adjustment to the self-attention layer of DiT block. 
Since both keyframes should be preserved while also synthesizing consistent intermediate frames, ReTRo increases the temporal RoPE~\cite{rope} scale near the keyframes and reduces it in other frames. 
The higher scale sharpens attention to maintain keyframe fidelity, while the lower scale broadens attention so intermediate frames attend more stably to both keyframes. 
Consequently, ReTRo reduces artifacts and blur, resulting in more temporally consistent GI.

Although text conditioning in GI enables more flexible and diverse synthesis, there is no reliable benchmark to evaluate models supporting text guidance.
We therefore introduce TGI-Bench, which curates sequences tailored for text-conditioned GI and pairs each with an aligned textual prompt.
Each sequence is annotated with one of four different \textit{challenge} categories in the GI field, enabling challenge-specific diagnosis of model strengths and weaknesses.

Our contributions are summarized as follows:
\begin{itemize}
    \item We propose KAB to deliver temporal and semantic guidance from the keyframes and text to the intermediate frames, improving semantic fidelity and pace stability.
    \item We also present ReTRo which rescales self-attention positional encodings to enhance overall frame consistency.
    \item We curate TGI-Bench for text-conditioned GI evaluation across sequence lengths and challenges, providing a diagnostic framework for future work.
\end{itemize}

\section{Related Work}
\label{section:relatedwork}
\begin{figure*}
  \centering
    \includegraphics[width=1\linewidth]{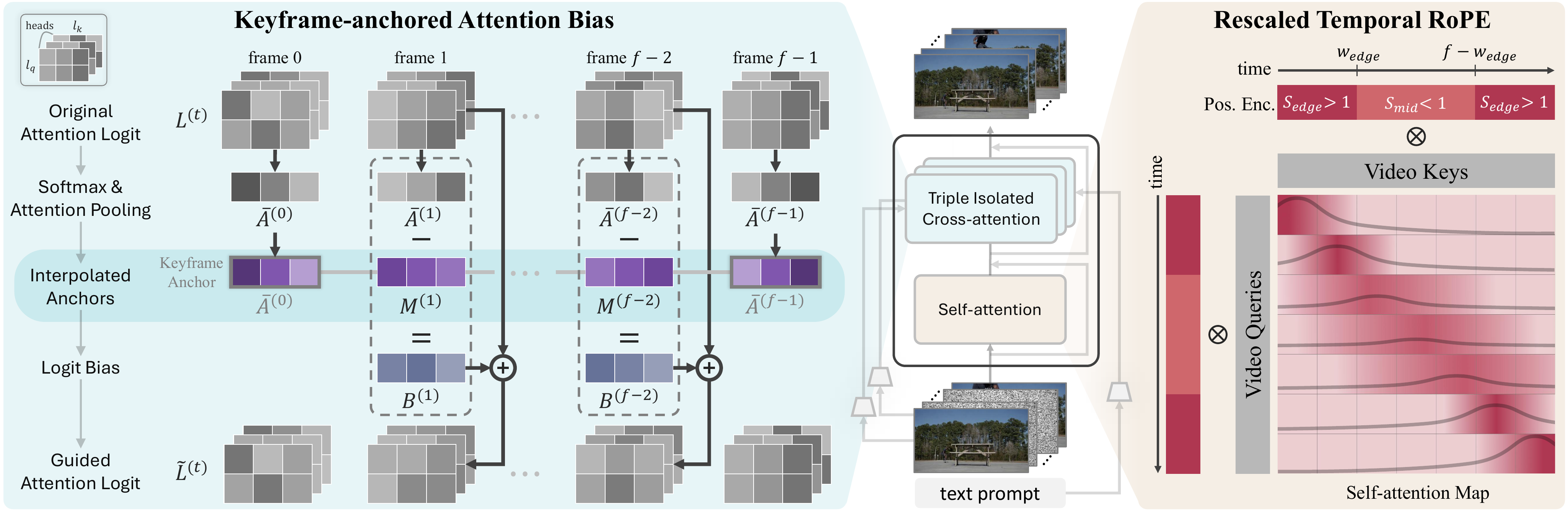}
    \caption{\textbf{Overall Pipeline of Our Method.} Our model is built upon a video DiT pipeline that consists of DiT blocks with self-attention and cross-attention layers. \textbf{Left:} \textit{Keyframe-anchored Attention Bias} is performed for each condition's cross-attention, which aggregates cross-attention maps from each keyframes to form keyframe anchors. These keyframe anchors are interpolated to frame-wise target anchors, which are used as a small logit bias to guide each intermediate frames.
    \textbf{Right:} Furthermore, we introduce \textit{Rescaled Temporal RoPE}, which increases temporal RoPE scale at the edges and reduces in the middle. As a result, edge frames place most of their attention on nearby frames while middle frames spread their attention across a wider temporal range. 
    }
    \vspace{-5pt}
    \label{fig:pipeline}
\end{figure*}
\paragraph{Generative Inbetweening.}
Traditional video frame interpolation (VFI) methods use deterministic pipelines, which are effective for small or near-linear motion but degrade under large displacements, non-linear dynamics, and occlusions \cite{reda2022film, ding2024video, cheng2021multiple}.
With the advent of diffusion models~\cite{ldm,svd,vdm}, large-scale training and sampling-based approaches emerged allowing VFI to expand to generative inbetweening (GI). 
For instance, TRF~\cite{trf} employs Stable Video Diffusion (SVD)~\cite{svd} in a bidirectional manner, and ViBiDSampler~\cite{vibid} advances it by altering the sampling strategy without additional training.
More studies such as Generative Inbetweening ~\cite{gi} and FCVG~\cite{fcvg} adopt feature-guided designs to improve motion consistency.
Meanwhile, Wan~\cite{wan} emerged as a video foundation model built on Diffusion Transformer backbone.
However, the attention mechanisms of DiT blocks optimized for the GI task remain underexplored.

\vspace{2pt}
\noindent\textbf{Cross-attention Editing.} 
Cross-attention has continuously been used as a control handle for text-driven image editing. 
Prompt-to-Prompt~\cite{prompt2prompt} preserves structure by copying and blending cross-attention maps across prompts and timesteps, while Attend-and-Excite~\cite{ane} reweights under-attended tokens to mitigate missing-object failures. 
Pix2Pix-Zero~\cite{pix2pixzero} keeps edits close to the source to maintain layout while changing semantics. 
Video-P2P~\cite{videop2p} extends this steering across frames to keep appearance consistent in video, and Layout Control~\cite{layoutctrl} manipulates cross-attention to satisfy user-specified boxes or landmarks.

\vspace{2pt}
\noindent\textbf{Rotary Positional Embeddings in Video DiT.}
RoPE scaling has been explored primarily in LLMs, showing that adjusting rotation rates can modulate locality and extend context with minimal changes~\cite{rope, peng2023yarn, scalinglaw}. 
Spatiotemporal RoPE designs stabilize long-range interactions but still rely on a uniform temporal schedule across frames~\cite{wei2025videorope}, and dynamic frequency schemes target diffusion steps rather than framewise control~\cite{issachar2025dype}. 
While cross-attention based positional schemes for temporal control exist~\cite{wu2025mind}, framewise temporal rescaling of RoPE inside self-attention for GI remains underexplored, to our knowledge.

\section{Method}
\label{section:method}
Our overall method pipeline is presented in ~\cref{fig:pipeline}.
We demonstrate our method on the First-Last-Frame-to-Video (FLF2V) pipeline in Wan2.1~\cite{wan}, a unique video DiT framework well-suited to our approach (\cref{method:pre}).
To ensure pace stability and semantic fidelity, we design target anchors which guide intermediate frames on the keyframes and text in the cross-attention (\cref{method:anchor}). 
Furthermore, we propose a scaling strategy within the self-attention that preserves consistency across frames (\cref{method:RoPE}).
Our method is model-agnostic and applies to any video DiT without additional training.

\subsection{Preliminary}
\label{method:pre}
To construct the video tokens, the video sequence of {\small$F$} frames is formed along the temporal axis by placing {\small$I_{\text{first}}$} at the beginning, {\small$I_{\text{last}}$} at the end, and inserting {\small$F-2$} zero-filled frames in between. 
This sequence is then compressed by Wan-VAE into a conditional latent sequence of $f$ frames.
After concatenating binary masks and latent diffusion noise with conditional latent, video tokens are obtained.

Meanwhile, to construct context vectors, the text prompt is encoded with UMT5 encoder~\cite{chung2023unimax} and projected to the DiT context space. 
For the two keyframes, CLIP~\cite{clip} features from {\small$I_{\text{first}}, I_{\text{last}}$} is concatenated and also projected to the context space. 
The final context vectors including all image and text embeddings are passed onto cross-attention layer.

\subsection{Keyframe-anchored Attention Bias}
\label{method:anchor}
We seek to guide the intermediate frames with semantic and temporal cues from the three conditions, two keyframes and text, by leveraging their cross-attention distributions.
As operating on full attention maps would be expensive, we compress these maps into \textit{keyframe anchors}, motivated by prior works~\cite{ane, masactrl,prompt2prompt,selfguidance}.
Interpolating between the two keyframe anchors yields frame-wise target anchors, which we use to add a small logit bias to each intermediate frame.

\vspace{2pt}
\noindent\textbf{Target Attention Anchor.}
Let {\small$L_h \coloneqq Q_h  K_h^{\top}/\sqrt{d_h}\!\in\!\mathbb{R}^{f l_q \times l_k}$} be the cross-attention logit for head {\small$h$}, where {\small$l_q$} and {\small$l_k$} are video query tokens per frame and condition key tokens, respectively.
Applying softmax, {\small$A_h \coloneqq \mathrm{softmax} \big( L_h \big)$} is a cross-attention heatmap whose rows are queries for {\small$f$} frames and columns are keys for each condition.
To build our anchors, we reuse the model's own cross-attention {\small$A_h$} and slice out the rows corresponding to each video frame to obtain {\small$A_h^{(t)} \in \mathbb{R}^{l_q \times l_k}$}.
These frame-wise slices are then compressed by averaging over heads and video queries to obtain frame-level distributions {\small$\bar{A}^{(t)} \in \mathbb{R}^{l_k}$}:
\begin{equation}
\small
\bar{A}^{(t)}=\mathop{\mathrm{Mean}}\limits_{H,l_q}\bigl[A_h^{(t)}\bigr], ~~~
t \in \{0,\ldots,f-1\}.
\end{equation}
We take {\small$\bar{A}^{(0)}$} and {\small$\bar{A}^{(f-1)}$} as the two keyframe anchors, capturing the keys that are globally important for the first and last frames of the video, respectively.

To guide each frame, we need a frame-wise target anchor {\small$M^{(t)} \in \mathbb{R}^{l_k}$} based on the two keyframe anchors.
For each intermediate frame index {\small$t \in \{1,\dots,f-2\}$}, we linearly interpolate the keyframe anchors to obtain:
\begin{equation}
\small
M^{(t)} \coloneqq (1-\tau^{(t)})\,\bar{A}^{(0)} + \tau^{(t)}\,\bar{A}^{(f-1)}, ~~~
\tau^{(t)} = \frac{t}{f-1}.
\end{equation}
These anchors can now be used to hint the model on the semantic and temporal information that should be emphasized at each frame for three conditions. 
By a slight abuse of notation, we also let $M^{(0)}=\bar{A}^{(0)}$ and $M^{(f-1)}=\bar{A}^{(f-1)}$.

\vspace{2pt}
\noindent\textbf{Frame-wise Logit Bias.}
Given the frame-wise target anchors, we gently steer the intermediate frames towards them by adding a small frame-wise logit bias {\small$B^{(t)} \in \mathbb{R}^{l_k}, t \in \{0,\ldots,f-1\}$} to the original cross-attention logits:
\begin{equation}
\small
\begin{aligned}
\widetilde L^{(t)}_h \coloneqq& L^{(t)}_h+\beta^{(t)} B^{(t)} \\
=& L^{(t)}_h+\beta^{(t)} \!\left(\log(M^{(t)}{+}\varepsilon)-\log(\bar A^{(t)}{+}\varepsilon)\right),
\end{aligned}
\end{equation}
    where {\small$B^{(t)}$} is added to all heads and all video queries of frame {\small$t$} (\ie, broadcasted), and {\small$\varepsilon$} is added to prevent near-zero possibilities.
Finally, we replace the original attention weights {\small$A_h$} with {\small$\tilde{A}_h=\mathrm{softmax}\big( \tilde{L}_h \big)$}.
This conservative \textit{pull} preserves the model’s local patterns while guiding its global token allocation toward the keyframes and text prompt. 

Following prior works, we also apply a smooth taper on {\small$\beta^{(t)}$} across the timeline, stronger near the keyframes and weaker in the middle, and gate the guidance to layers $5$--$12$ only and step $1$ to \(40\%\) of the total steps~\cite{ldm,ane,prompt2prompt}.
This nudges the semantic and temporal guidance to settle early while leaving late steps and layers to form textures and fine details, since mid-level layers often carry much of the image’s spatial or semantic structure \cite{understandingediting}.

\vspace{2pt}
\noindent\textbf{Triple Isolated Cross-attention.}
In the baseline FLF2V pipeline, the model applies cross-attention to {\small$I_{\text{first}}$} alone, while {\small$I_{\text{last}}$} and {\small$\mathrm{text}$} are concatenated and attended jointly, yielding an asymmetric fusion across the three conditions. 
Unlike the baseline’s asymmetric fusion, we compute three symmetric cross-attentions separately: {\small$I_{\text{first}}\leftrightarrow\mathrm{video}$}, {\small$I_{\text{last}}\leftrightarrow\mathrm{video}$}, and {\small$\mathrm{text}\leftrightarrow\mathrm{video}$}.
Each cross-attention is refined by its own {\small$M^{(t)}$} and {\small$\beta^{(t)}$}, and then equally weighted and averaged, amplifying the effect of explicit semantic guidance while preserving symmetry across modalities.

\begin{figure*}
  \centering
    \includegraphics[width=1\linewidth]{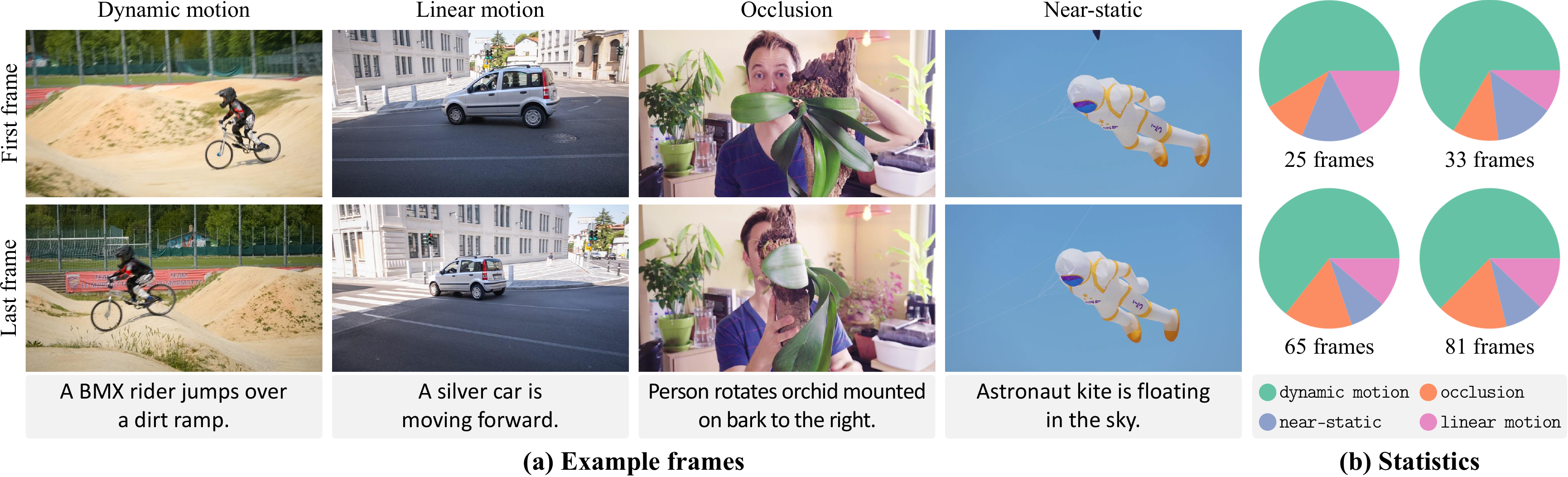}
    \caption{
    \textbf{TGI-Bench.}
      \textbf{(a)} One example from each challenge of our TGI-Bench is presented. For each example, the first and last frames of the video along with its text description are shown. \textbf{(b)} The distribution of challenges according to the number of frames is illustrated.
      }
      \vspace{-8pt}
    \label{fig:data_stats}
\end{figure*}

\subsection{Rescaled Temporal RoPE}
\label{method:RoPE}
Prior video DiT models employ RoPE within self-attention to inject relative spatiotemporal information by rotating queries and keys in phase. 
However, generative inbetweening task imposes a different challenge that two keyframes must be preserved and intermediate frames should be generated while jointly conditioning on both keyframes.
Vanilla RoPE provides relative distances of frames, but lacks explicit mechanism that anchors the two keyframes, which leads to frame inconsistency. 

We therefore introduce Rescaled Temporal RoPE (ReTRo), inspired by RoPE scaling methods for LLMs \cite{chen2023extending, peng2023yarn}.
Specifically, we apply higher RoPE scales near the two keyframes and lower scales in other frames, sharpening locality to preserve the keyframes while broadening attention to promote consistency across the intermediate frames.

First, we construct per-axis frequency rows (temporal, height, width) and concatenate them: 
\begin{equation}
\small
\Psi(t,h,w) \coloneqq \big[\Omega_t[t];\Omega_h[h];\Omega_w[w]\big], ~~~
t\in\{0,\ldots,f{-}1\}.
\end{equation}
Then, we pick an integer {\small$w_{\text{edge}}\!\in\!\{0,\ldots,\lfloor f/2\rfloor\}$}, which stands for number of edge frames per side, and define the edge and middle index sets symmetrically:
\begin{equation}
\small
\begin{aligned}
\mathcal{T}_{\text{edge}} & \coloneqq \{0,\ldots,w_{\text{edge}}-1\}\\
&\quad\cup\,\{f-w_{\text{edge}},\ldots,f-1\},\\
\mathcal{T}_{\text{mid}}  & \coloneqq \{w_{\text{edge}},\ldots,f-w_{\text{edge}}-1\}.
\end{aligned}
\end{equation}
We scale the temporal frequency row per frame to maintain edge frames fidelity and stabilize the middle:

\begin{equation}
\small
s(t) \coloneqq
\begin{cases}
s_{\text{edge}}, & t \in \mathcal{T}_{\text{edge}},\\
s_{\text{mid}},  & t \in \mathcal{T}_{\text{mid}},
\end{cases}
\qquad
\tilde{\Omega}_t[t] = s(t)\,\Omega_t[t],
\end{equation}
where $s_{\text{edge}} > 1$ and $s_{\text{mid}} < 1$.

Finally, we reassemble the per-axis frequency rows and use it for both video queries and keys:
\begin{equation}
\small
\Psi_{\text{ReTRo}}(t,h,w) \coloneqq \big[\tilde\Omega_t[t];\Omega_h[h];\Omega_w[w]\big], \end{equation} \begin{equation} \small \tilde Q=\mathrm{RoPE}(Q;\Psi_{\text{ReTRo}}), ~~ \tilde K=\mathrm{RoPE}(K;\Psi_{\text{ReTRo}}). \end{equation}

As a result, in self-attention the edge frames place most of their attention on nearby frames, stabilizing local detail and keyframe fidelity. In contrast, the middle frames spread their attention across a wider temporal range, drawing information from more distant frames to enhance overall consistency. 
This yields a simple, training-free mechanism for frame consistency without architectural changes.
\section{TGI-Bench}
\label{section:dataset}
Previous studies on generative inbetweening (GI) and video frame interpolation have primarily relied on video datasets such as \cite{davis, lavib} for model evaluation.
However, while these benchmarks provide dense ground-truth frames, most of these resources lack natural-language annotations, restricting the ability to evaluate whether a model truly reflects textual instructions when generating videos.
In addition, the existing benchmarks do not offer diverse challenges, which hinders diagnosing a model across various capabilities.

Thus, we present Text-conditioned Generative Inbetweening Benchmark (TGI-Bench). 
TGI-Bench consists of two keyframes, the ground-truth intermediate frames, a textual description, and its designated challenge category.
To cover both short and long sequences, we release four sequence-length variants for 25, 33, 65, and 81 frames, following prior works~\cite{svd, opensora, nova, emu, wan}, supporting broad, apples-to-apples comparison across methods.
In summary, TGI-Bench (i) enables the evaluation of a model's text-grounded generative inbetweening performance (ii) across diverse sequence lengths, (iii) while also allowing for the fine-grained diagnosis of its capabilities within specific challenge categories.

\vspace{2pt}
\noindent\textbf{Data Curation.}
To construct TGI-Bench dataset, we select videos from the DAVIS~\cite{davis} dataset, as well as from the Pexels and Pixabay websites\footnote{https://www.pexels.com/, https://www.pixabay.com/}. 
Videos without a clear main object or with overly complex motion that cannot be sufficiently described with text are excluded, resulting in a final selection of 220 videos.
From each video, we uniformly sample {\small$F$} frames. 
These {\small$F$} frames are then subsampled and provided to GPT-4.1~\cite{gpt4}, which generates a text description and assigns one of the following challenge types~\cite{ace}: \texttt{dynamic motion, occlusion, linear motion, near-static}.
We repeat this process for {\small$F \in \{25, 33, 65, 81\}$}, resulting in four validation sets corresponding to different video lengths.
The generated descriptions and challenge labels are then manually reviewed and corrected by the authors to ensure accuracy.
\cref{fig:data_stats}(a) shows one example per challenge for {\small$F = 25$}, and \cref{fig:data_stats}(b) shows the challenge distribution for different values of {\small$F$}.
Please refer to the supplementary material for the detailed GPT prompt and sampling process.

\section{Experiments}
\label{section:experiments}
\setlength{\tabcolsep}{2pt} 
\renewcommand{\arraystretch}{1} 
\setlength{\aboverulesep}{0.5pt}
\setlength{\belowrulesep}{0.5pt}
\setlength{\abovetopsep}{0pt}
\setlength{\belowbottomsep}{0pt}
\setlength{\cmidrulekern}{0pt} 

\begin{table*}[t]
\caption{\textbf{Video Generation Evaluation Results. } Quantitative comparison of the baselines and our method on 65, 81 frames. We evaluate video generation quality and fidelity. The best results are in \textbf{bold}, and the second best are \underline{underlined.}} 
\vspace{-5pt}
\centering
\small
\begin{tabular*}{\textwidth}{@{\extracolsep{\fill}} @{} l *{12}{c} @{}}
\toprule
\multirow{2}{*}{Method} &
\multicolumn{6}{c}{\textbf{65-frame}} &
\multicolumn{6}{c}{\textbf{81-frame}} \\
\cmidrule(l{0.2em}r{0.2em}){2-7}\cmidrule(l{0.2em}r{0.2em}){8-13}
& PSNR$\uparrow$ & SSIM$\uparrow$ & LPIPS$\downarrow$ & FID$\downarrow$ & FVD$\downarrow$
& VBench$\uparrow$
& PSNR$\uparrow$ & SSIM$\uparrow$
& LPIPS$\downarrow$ & FID$\downarrow$ & FVD$\downarrow$
& VBench$\uparrow$ \\
\midrule
TRF~\cite{trf} & 16.06 & \underline{0.5662} & 0.5173 & 168.37 & 0.3379 & 8.117
& 16.08 & 0.5859 & 0.5136 & 169.46 & 0.3324 & 8.074 \\

ViBiDSampler~\cite{vibid} & 15.45 & 0.5381 & 0.5211 & 160.74 & 0.3574 & 8.619
& 15.57 & 0.5533 & 0.5142 & 163.09 & 0.3482 & 8.759 \\

GI~\cite{gi} & 15.43 & 0.5473 & 0.5210 & 224.36 & 0.3693 & 7.971
& 15.59 & 0.5666 & 0.5166 & 219.80 & 0.3416 & 7.923 \\

FCVG~\cite{fcvg} & \underline{16.77} & 0.5412 & 0.4412 & 98.02 & \underline{0.2938} & 9.755 
& 17.16 & 0.5698 & 0.4216 & 97.01 & \underline{0.2607} & 9.899 \\

Wan~\cite{wan} & 16.75 & 0.5661 & \underline{0.4172} & \underline{82.42} & 0.3406 & \underline{9.861}
& \underline{17.63} & \underline{0.6179} & \underline{0.3945} & \underline{82.90} & 0.2769 & \underline{9.904} \\
\hline
\textbf{Ours} & \textbf{17.68} & \textbf{0.5903} & \textbf{0.4016} & \textbf{77.66} & \textbf{0.2820} & \textbf{9.924} 
& \textbf{18.17} & \textbf{0.6269} & \textbf{0.3818} & \textbf{77.59} & \textbf{0.2458} & \textbf{10.022} \\
\bottomrule
\end{tabular*}
\label{tab:quan1}
\end{table*}

\setlength{\tabcolsep}{2pt} 
\renewcommand{\arraystretch}{1} 
\setlength{\aboverulesep}{0.5pt}
\setlength{\belowrulesep}{0.5pt}
\setlength{\abovetopsep}{0pt}
\setlength{\belowbottomsep}{0pt}
\setlength{\cmidrulekern}{0pt} 

\begin{table*}[t]
\caption{{\textbf{Generative Inbetweening Evaluation Results.} To complement the video generation evaluation, we conduct experiments on other metrics that reflect our target qualities. L-\textit{frames} stands for LPIPS-\textit{frames} and C-\textit{frames} stands for CLIPSIM-\textit{frames}. For human evaluation, we evaluate FC, SF, PS which stands for frame consistency, semantic fidelity and pace stability, respectively. The best results are in \textbf{bold}, and the second best are \underline{underlined}. }}
\vspace{-5pt}
\centering
\small
\begin{tabular*}{\textwidth}{@{\extracolsep{\fill}} @{}l cc cc cc cc ccc@{}} 
\toprule
\multirow{3}{*}{Method} &
\multicolumn{4}{c}{\textbf{65-frame}} &
\multicolumn{7}{c}{\textbf{81-frame}} \\
\cmidrule(l{0.2em}r{0.2em}){2-5}\cmidrule(l{0.2em}r{0.2em}){6-12}
& \multicolumn{2}{c}{Sem. Fid.}   &
  \multicolumn{2}{c}{Frame Cons.} &
  \multicolumn{2}{c}{Sem. Fid.}   &
  \multicolumn{2}{c}{Frame Cons.} &
  \multicolumn{3}{c}{Human Eval.}  \\
\cmidrule(l{0.2em}r{0.2em}){2-3}\cmidrule(l{0.2em}r{0.2em}){4-5}
\cmidrule(l{0.2em}r{0.2em}){6-7}\cmidrule(l{0.2em}r{0.2em}){8-9}\cmidrule(l{0.2em}r{0.2em}){10-12}
& X\mbox{-}CLIP$\uparrow$ & VQA$\uparrow$ & L-\textit{frames}$\downarrow$ & C-\textit{frames}$\uparrow$
& X\mbox{-}CLIP$\uparrow$ & VQA$\uparrow$ & L-\textit{frames}$\downarrow$ & C-\textit{frames}$\uparrow$ & FC$\uparrow$& SF$\uparrow$ & PS$\uparrow$ \\
\midrule

TRF~\cite{trf} 
& 0.2174 & 0.5720 & 0.1869 & 0.9669  
& 0.2089 & 0.5465 & 0.1883 & 0.9647 & 1.60 & 1.53 & 2.10  \\

ViBiDSampler~\cite{vibid} 
& 0.2249 & 0.4982 & 0.1782 & 0.9722  
& 0.2186 & 0.5396 & 0.1751 & 0.9733 & 2.05 & 1.82 & 2.38 \\

GI~\cite{gi} 
& 0.2169 & 0.4901 & 0.1632 & 0.9742 
& 0.2082 & 0.4545 & 0.1692 & 0.9735 & 1.70 & 1.55 & 1.89  \\

FCVG~\cite{fcvg} 
& 0.2256 & 0.6338 & \textbf{0.0792 }& \underline{0.9831}
& 0.2222 & 0.6194 & \textbf{0.0613} & \underline{0.9858} & 2.87 & 2.73 & 2.91  \\

Wan~\cite{wan} 
& \underline{0.2326} & \underline{0.6631} & 0.1334 & 0.9788
& \underline{0.2262} & \underline{0.6730} & 0.0959 & 0.9839 & \underline{3.50} & \underline{3.69} & \underline{3.65} \\
\hline
\textbf{Ours} 
& \textbf{0.2340} & \textbf{0.6692} & \underline{0.1047} & \textbf{0.9855}
& \textbf{0.2292} & \textbf{0.6752} & \underline{0.0774} & \textbf{0.9881} & \textbf{4.38} & \textbf{4.27} & \textbf{4.34}  \\
\bottomrule
\vspace{-10pt}
\end{tabular*}
\label{tab:quan2}
\end{table*}


\noindent\textbf{Baselines \& Dataset.}
We compare our model with baselines including TRF~\cite{trf}, ViBiDSampler~\cite{vibid}, GI~\cite{gi}, FCVG~\cite{fcvg} and Wan2.1~\cite{wan}. 
For all baselines and our method, we use our TGI-Bench to evaluate the generated video on various metrics shown in \cref{sec:quan}. 
Each method is assessed under four frame-length settings, following the structure defined in our TGI-Bench.
Additional examples are deferred to the supplementary material.

\vspace{2pt}
\noindent\textbf{Implementation Details.}
For strict comparison, the seed is fixed across all methods and experiment.
When implementing Keyframe-anchored Attention Bias (KAB), the attention logit bias strength {\small$\beta^{(i)}$} uses a cosine taper, taking the value $0.7$ near the keyframes and decreasing toward the midpoint to $0.3$.
For Rescaled Temporal RoPE (ReTRo), the width of edge frames {\small$w_{\text{edge}}$} is defined as an integer closest to 0.1 times the total number of frames. We fix the scaling factors {\small$s_{\text{edge}}=1.06$} and {\small$s_{\text{mid}}=0.94$}.


\subsection{Quantitative Evaluation}
\label{sec:quan}
\noindent\textbf{Video Generation Evaluation.}
Due to page limit, we report the 65- and 81-frame results as longer horizons are more discriminative and practically relevant.
We compute PSNR, SSIM~\cite{ssim}, and LPIPS~\cite{lpips} against ground truth frames, use FVD~\cite{fvd}, FID~\cite{fid} and VBench~\cite{vbench} score to measure overall video quality.
We select 8 relevant dimensions for VBench: I2V Subject, I2V Background, Subject Consistency, Background Consistency, Motion Smoothness, Aesthetic Quality and Imaging Quality.
For all metrics, our method achieves the best performance as shown in \cref{tab:quan1}.
Further details and complete results on shorter sequences are in supplementary material.

\vspace{2pt}
\noindent\textbf{Generative Inbetweening Evaluation.} 
While traditional video generation metrics are effective at measuring frame consistency, they are insufficient for semantic fidelity and pace stability. 
Thus, we additionally assess these qualities to evaluate our method.
Semantic fidelity is measured by X\mbox{-}CLIP~\cite{xclip} text-to-video similarity between prompt and generated video embeddings. 
For the video visual question answering (VQA) score, we average over 6 different QA models to reduce variance.
Frame consistency is further evaluated with LPIPS-\textit{frames} and CLIPSIM-\textit{frames}, which average similarities between adjacent frames following prior works~\cite{storydiff, wave}. 
Interestingly FCVG, which provides intermediate-frame motion guidance that likely reduces path ambiguity, yields comparable performance to text-conditioned GI without any text input.
As automatic measurements are unreliable to properly evaluate our target qualities, semantic fidelity and pace stability, we run a user study on 10\% of the video samples, focusing on 81-frame videos.
As shown in \cref{tab:quan2}, our method improves both semantic fidelity and frame consistency. 
It also attains the highest human-rated semantic fidelity and pace stability.
More results and detailed protocols for the VQA score and user study are provided in the supplementary material.

\subsection{Qualitative Evaluation}
\begin{figure*}[!]
  \centering
    \includegraphics[width=\linewidth]{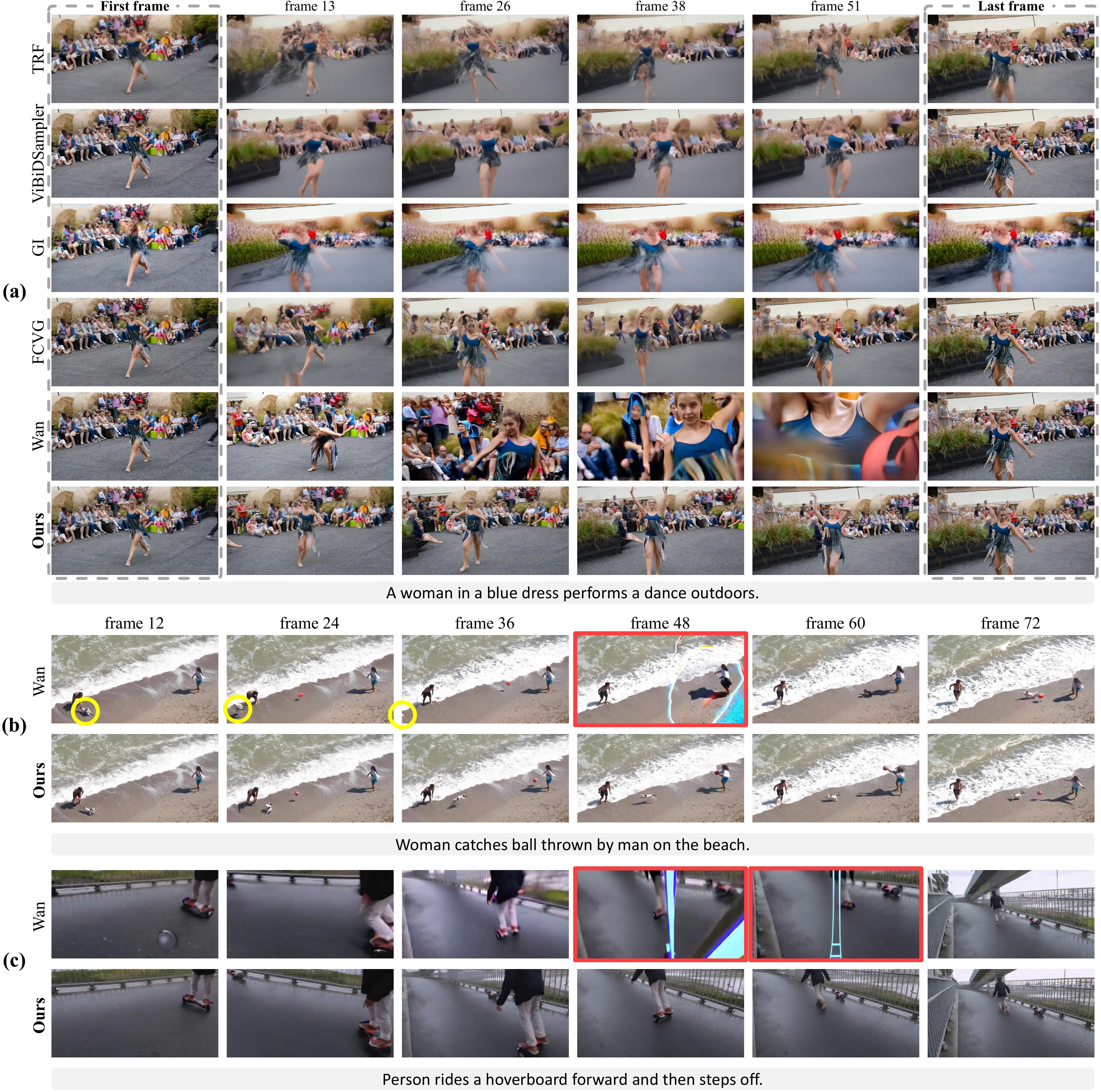}
    \caption{
      \textbf{Qualitative Comparison with Baselines.} \textbf{(a)} Our method outperforms prior works in all three target challenges: semantic fidelity, pace stability and frame consistency. Although Wan performs better than SVD-based models such as TRF, ViBiDSampler, GI and FCVG, it shows failures in one or more qualities. 
      For instance, for \textbf{(b)}, the dog marked by a yellow circle (\yellow{\faCircleO}) disappears to the left and suddenly reappears in the middle of the frame in later sequences, showing semantic infidelity while in \textbf{(c)}, the location of the person barely moves from frame 48 to frame 60, showing pace instability. On the other hand, our method overcomes all three challenges. 
      }
      \vspace{-8pt}
    \label{fig:qual1}
\end{figure*}


The visual comparison in ~\cref{fig:qual1} shows a representative qualitative comparison of the baselines and our method. 
Prior to any analysis of semantic fidelity or pace stability, we already observe visible artifacts such as object collapse, uneven motion, and blurred backgrounds in the SVD-based methods such as TRF, ViBiDSampler, GI, and FCVG.
Although Wan preserves frame consistency, it does not follow the prompt. 
The prompt requires rotating the mallet counterclockwise, Wan repeatedly moves the mallet up and down with an uneven pace. 
In contrast, our method generates a smooth counterclockwise rotation at a natural pace that matches the prompt.
Please refer to the supplementary material for more qualitative results.

\subsection{Ablation Study}
We conduct an ablation study in ~\cref{tab:ablation} on our two methods, Keyframe-anchored Attention Bias (KAB) and Rescaled Temporal RoPE (ReTRo).
Across video generation metrics, our method achieves the highest scores on most measures, with the ReTRo-only variant as second-best overall. 

ReTRo strengthens temporal cues within each intermediate frame and pulls self-attention toward both keyframes. 
This temporal reference is particularly well suited to improving frame consistency, which aligns with the competitive quantitative scores of the ReTRo-only variant. 
Contrastly, KAB operates in the cross-attention layer, allowing guidance from keyframe anchors gained from the keyframes as well as the text prompt, applying a lightweight per-token logit bias for each intermediate frame. 
This mechanism supplies semantic and temporal guidance, making it especially effective for pace stability and semantic fidelity.

\begin{table}[]
\centering
\renewcommand{\arraystretch}{1}
\small
\caption{\textbf{Ablation Results on KAB and ReTRo.} Applying both methods achieves the best scores most number of metrics. ReTRo is the next best overall, reflecting its strength on frame consistency.}
\resizebox{\linewidth}{!}{%
    \begin{tabular}{ccccccccc}
    \toprule
    {\footnotesize Exp.} & \multirow{2}{*}{KAB} & \multirow{2}{*}{ReTRo} &
    \multicolumn{6}{c}{81-frame} \\
    \cline{4-9}
    \# & & &
    PSNR$\uparrow$ & SSIM$\uparrow$ & LPIPS$\downarrow$ &
    FID$\downarrow$ & FVD$\downarrow$ & VBench$\uparrow$ \\
    \hline
    1 & & &
    17.63 & 0.6179 & 0.3945 & 82.90 & 0.2769 & \underline{9.904} \\

    2 & \checkmark & &
    17.62 & 0.6174 & 0.3945 & 83.01 & 0.2738 & 9.896 \\

    3 & & \checkmark &
    \textbf{18.18} & \textbf{0.6309} & \underline{0.3892} &
    \underline{77.88} & \underline{0.2510} & 9.749 \\

    \hline
    4 & \checkmark & \checkmark &
    \underline{18.17} & \underline{0.6269} & \textbf{0.3818} &
    \textbf{77.59} & \textbf{0.2458} & \textbf{10.022} \\
    \bottomrule
    \end{tabular}
}
\vspace{-8pt}
\label{tab:ablation}
\end{table}
\begin{figure*}[t]
  \centering
    \includegraphics[width=\linewidth]{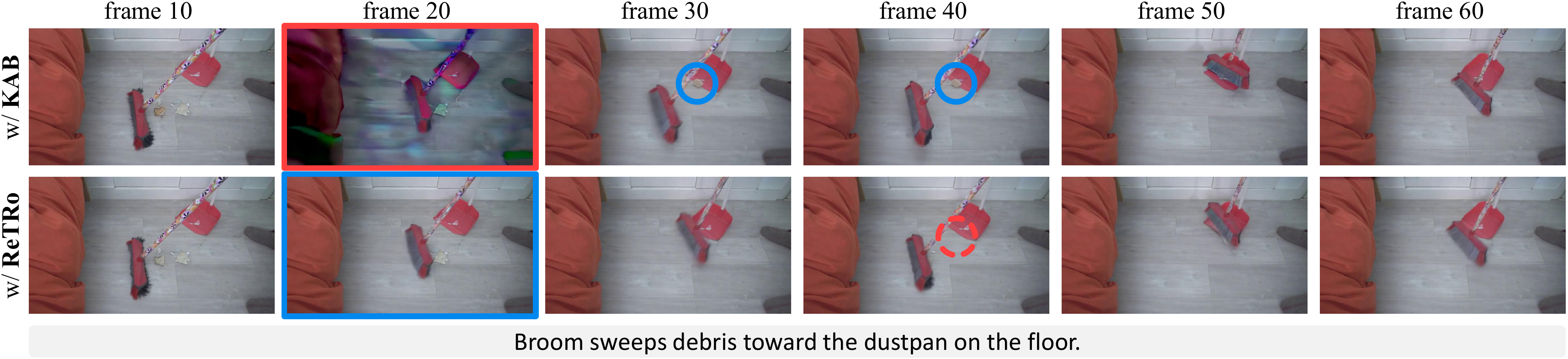}
    \caption{
      \textbf{Qualitative Comparison of KAB and ReTRo.} This figure shows a qualitative example of the different roles of KAB and ReTRo. KAB preserves the semantically important trash object (\blue{\faCircleO}) consistently across frames 30–40, whereas ReTRo suppresses artifacts around frame 20. Together they address different failure modes, indicating that KAB and ReTRo are complementary rather than interchangeable.
      } 
      \vspace{-7pt}
    \label{fig:ablation}
\end{figure*}

Since these effects difficult to measure faithfully by standard video generation metrics, we also include qualitative results shown in \cref{fig:ablation}.
For instance, KAB preserves a semantically important object consistently across frames 30 to 40, while ReTRo removes the artifact visible around frame 20. 
Together, these complementary effects yield more consistent and semantically faithful GI with natural motions.

\subsection{Generative Inbetweening Challenge Analysis}
\begin{figure}
\centering 
\includegraphics[width=0.98\columnwidth]{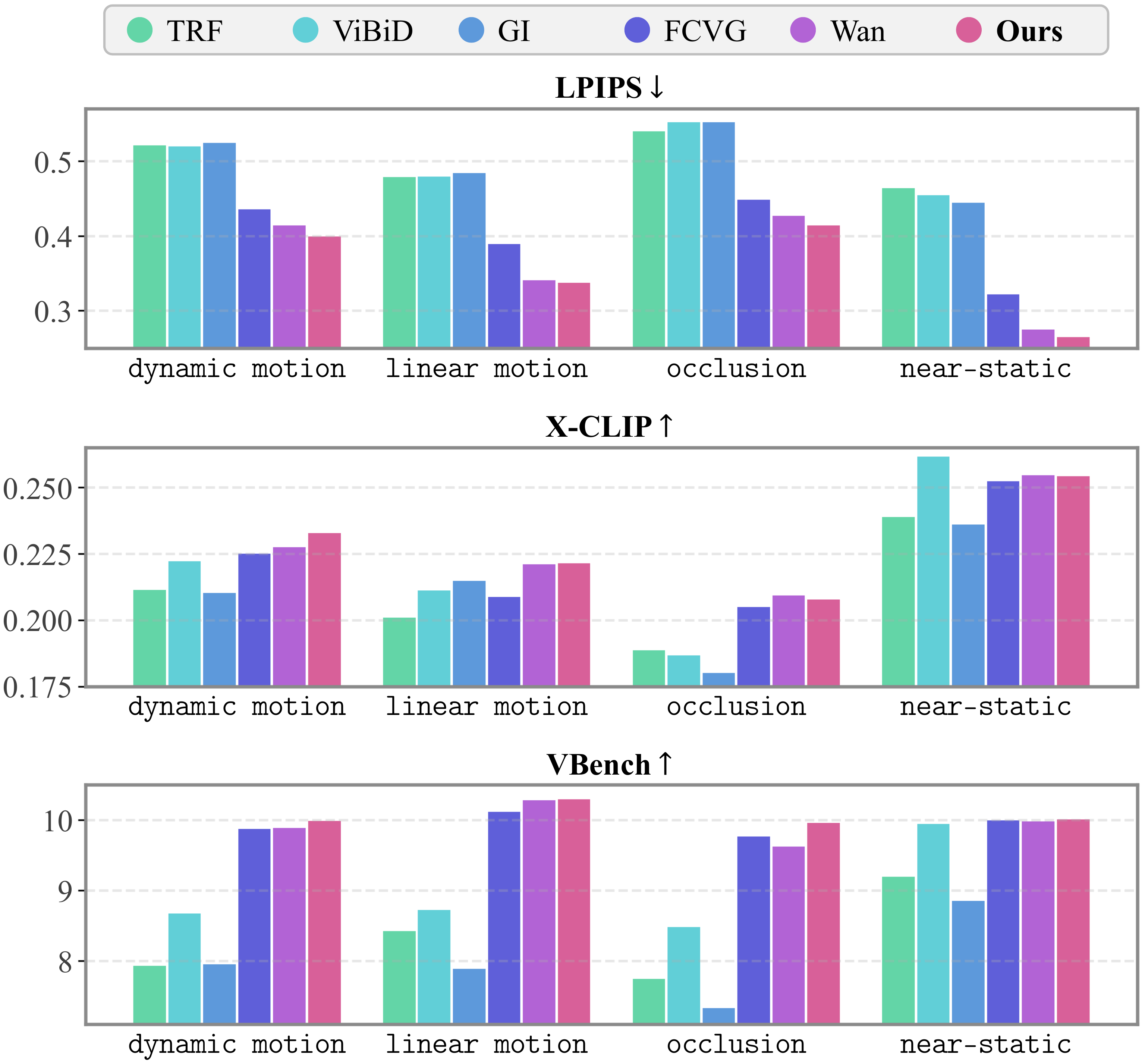} 
\vspace{-3pt}
    \caption{
    \textbf{Generative Inbetweening Challenge Analysis.} We utilize our TGI-Bench to diagnose existing GI mothods across four challenges. This reveals a clear difficulty spectrum: most methods do well on \texttt{near-static}, while \texttt{occlusion} is the hardest.
    } 
    \vspace{-14pt}
    \label{fig:challenge} 
\end{figure}

To evaluate whether TGI-Bench supports fine-grained diagnosis of model strengths and weaknesses, we assess existing GI methods across the four different challenges: \texttt{dynamic motion, linear motion, occlusion, near-static}.
As shown in ~\cref{fig:challenge}, our benchmark reveals a clear spectrum of difficulty. 
While most methods perform reasonably well on \texttt{near-static}, \texttt{occlusion} emerges as the most difficult for all models.

Across all four challenges, our method compares favorably to existing GI models on the key metrics, LPIPS, X-CLIP, and VBench. 
This advantage is especially clear on the persistent GI challenges of \texttt{occlusion} and \texttt{dynamic motion}, where our method achieves the best LPIPS and VBench scores, indicating improved frame consistency.

We additionally analyze how text conditioning and robustness to challenging cases affect overall evaluation.
Methods that do not accept textual input, such as TRF, ViBiDSampler and GI, lag behind text-conditioned methods Wan and our method for harder challenges. 
Although FCVG is not text-conditioned, it leverages motion guidance and delivers respectable performance across individual challenge categories.
When compared against human evaluation shown in ~\cref{tab:quan2}, we also observe that differences on the harder subsets account for most variation in human scores. 
Methods that better handle \texttt{dynamic motion} and \texttt{occlusion} are consistently preferred, making these challenging subsets a more informative indicator of human-perceived quality than \texttt{near-static} cases alone.

\section{Conclusion}
\label{section:conclusion}

In this paper, we propose two training-free mechanisms for generative inbetweening that can act as a general plug-in to any video DiT models.
First, Keyframe-anchored Attention Bias (KAB) utilizes the model’s own cross-attention layer on the first and last frames as signals and guide intermediate frames toward a per-frame interpolation of these keyframe-anchors.
Furthermore, we incorporate Rescaled Temporal RoPE (ReTRo) which enlarges temporal RoPE scale at the edge frames and reduces in the middle to enhance frame consistency.
Finally, we release TGI-Bench, a benchmark that systematically diagnoses the current challenges of generative inbetweening. 
We expect this benchmark to accelerate the field of text-conditioned GI, especially under dynamic motion and occlusion.

\paragraph{Acknowledgement.}
This work was supported in part by the IITP RS-2024-00457882 (AI Research Hub Project), IITP 2020-II201361, NRF RS-2024-00345806, NRF RS-2023-002620, and RQT-25-120390. Affiliations: Department of Artificial Intelligence (T.E.C, J.K, S.J.H), Department of Computer Science (S.S).

{
    \small
    \bibliographystyle{ieeenat_fullname}
    \bibliography{main}

\begin{thebibliography}{52}
\providecommand{\natexlab}[1]{#1}
\providecommand{\url}[1]{\texttt{#1}}
\expandafter\ifx\csname urlstyle\endcsname\relax
  \providecommand{\doi}[1]{doi: #1}\else
  \providecommand{\doi}{doi: \begingroup \urlstyle{rm}\Url}\fi

\bibitem[Blattmann et~al.(2023)Blattmann, Dockhorn, Kulal, Mendelevitch, Kilian, Lorenz, Levi, English, Voleti, Letts, Jampani, and Rombach]{svd}
Andreas Blattmann, Tim Dockhorn, Sumith Kulal, Daniel Mendelevitch, Maciej Kilian, Dominik Lorenz, Yam Levi, Zion English, Vikram Voleti, Adam Letts, Varun Jampani, and Robin Rombach.
\newblock Stable video diffusion: Scaling latent video diffusion models to large datasets, 2023.

\bibitem[Cao et~al.(2023)Cao, Wang, Qi, Shan, Qie, and Zheng]{masactrl}
Mingdeng Cao, Xintao Wang, Zhongang Qi, Ying Shan, Xiaohu Qie, and Yinqiang Zheng.
\newblock Masactrl: Tuning-free mutual self-attention control for consistent image synthesis and editing, 2023.

\bibitem[Chefer et~al.(2023)Chefer, Alaluf, Vinker, Wolf, and Cohen-Or]{ane}
Hila Chefer, Yuval Alaluf, Yael Vinker, Lior Wolf, and Daniel Cohen-Or.
\newblock Attend-and-excite: Attention-based semantic guidance for text-to-image diffusion models, 2023.

\bibitem[Chen et~al.(2023{\natexlab{a}})Chen, Laina, and Vedaldi]{layoutctrl}
Minghao Chen, Iro Laina, and Andrea Vedaldi.
\newblock Training-free layout control with cross-attention guidance, 2023{\natexlab{a}}.

\bibitem[Chen et~al.(2023{\natexlab{b}})Chen, Wong, Chen, and Tian]{chen2023extending}
Shouyuan Chen, Sherman Wong, Liangjian Chen, and Yuandong Tian.
\newblock Extending context window of large language models via positional interpolation.
\newblock \emph{arXiv preprint arXiv:2306.15595}, 2023{\natexlab{b}}.

\bibitem[Cheng and Chen(2021)]{cheng2021multiple}
Xianhang Cheng and Zhenzhong Chen.
\newblock Multiple video frame interpolation via enhanced deformable separable convolution.
\newblock \emph{IEEE Transactions on Pattern Analysis and Machine Intelligence}, 44\penalty0 (10):\penalty0 7029--7045, 2021.

\bibitem[Chung et~al.(2023)Chung, Constant, Garcia, Roberts, Tay, Narang, and Firat]{chung2023unimax}
Hyung~Won Chung, Noah Constant, Xavier Garcia, Adam Roberts, Yi Tay, Sharan Narang, and Orhan Firat.
\newblock Unimax: Fairer and more effective language sampling for large-scale multilingual pretraining.
\newblock \emph{arXiv preprint arXiv:2304.09151}, 2023.

\bibitem[Deng et~al.(2024)Deng, Pan, Diao, Luo, Cui, Lu, Shan, Qi, and Wang]{nova}
Haoge Deng, Ting Pan, Haiwen Diao, Zhengxiong Luo, Yufeng Cui, Huchuan Lu, Shiguang Shan, Yonggang Qi, and Xinlong Wang.
\newblock Autoregressive video generation without vector quantization.
\newblock \emph{arXiv preprint arXiv:2412.14169}, 2024.

\bibitem[Ding et~al.(2024)Ding, Lin, Zhang, Liu, and Yu]{ding2024video}
Chao Ding, Mingyuan Lin, Haijian Zhang, Jianzhuang Liu, and Lei Yu.
\newblock Video frame interpolation with stereo event and intensity cameras.
\newblock \emph{IEEE Transactions on Multimedia}, 26:\penalty0 9187--9202, 2024.

\bibitem[Elhage et~al.(2021)Elhage, Nanda, Olsson, Henighan, Joseph, Mann, Askell, Bai, Chen, Conerly, et~al.]{transformer_circuit}
Nelson Elhage, Neel Nanda, Catherine Olsson, Tom Henighan, Nicholas Joseph, Ben Mann, Amanda Askell, Yuntao Bai, Anna Chen, Tom Conerly, et~al.
\newblock A mathematical framework for transformer circuits.
\newblock \emph{Transformer Circuits Thread}, 1\penalty0 (1):\penalty0 12, 2021.

\bibitem[Epstein et~al.(2023)Epstein, Jabri, Poole, Efros, and Holynski]{selfguidance}
Dave Epstein, Allan Jabri, Ben Poole, Alexei~A. Efros, and Aleksander Holynski.
\newblock Diffusion self-guidance for controllable image generation, 2023.

\bibitem[et~al.(2024)]{gpt4}
OpenAI et al.
\newblock Gpt-4 technical report, 2024.

\bibitem[et~al.(2025)]{wan}
Team~Wan et al.
\newblock Wan: Open and advanced large-scale video generative models, 2025.

\bibitem[Feng et~al.(2024)Feng, Ding, Xia, Niklaus, Abrevaya, Black, and Zhang]{trf}
Haiwen Feng, Zheng Ding, Zhihao Xia, Simon Niklaus, Victoria Abrevaya, Michael~J. Black, and Xuaner Zhang.
\newblock Explorative inbetweening of time and space, 2024.

\bibitem[Girdhar et~al.(2023)Girdhar, Singh, Brown, Duval, Azadi, Rambhatla, Shah, Yin, Parikh, and Misra]{emu}
Rohit Girdhar, Mannat Singh, Andrew Brown, Quentin Duval, Samaneh Azadi, Sai~Saketh Rambhatla, Akbar Shah, Xi Yin, Devi Parikh, and Ishan Misra.
\newblock Emu video: Factorizing text-to-video generation by explicit image conditioning.
\newblock \emph{arXiv preprint arXiv:2311.10709}, 2023.

\bibitem[Hertz et~al.(2022)Hertz, Mokady, Tenenbaum, Aberman, Pritch, and Cohen-Or]{prompt2prompt}
Amir Hertz, Ron Mokady, Jay Tenenbaum, Kfir Aberman, Yael Pritch, and Daniel Cohen-Or.
\newblock Prompt-to-prompt image editing with cross attention control, 2022.

\bibitem[Heusel et~al.(2017)Heusel, Ramsauer, Unterthiner, Nessler, and Hochreiter]{fid}
Martin Heusel, Hubert Ramsauer, Thomas Unterthiner, Bernhard Nessler, and Sepp Hochreiter.
\newblock Gans trained by a two time-scale update rule converge to a local nash equilibrium.
\newblock \emph{Advances in neural information processing systems}, 30, 2017.

\bibitem[Ho et~al.(2022)Ho, Salimans, Gritsenko, Chan, Norouzi, and Fleet]{vdm}
Jonathan Ho, Tim Salimans, Alexey Gritsenko, William Chan, Mohammad Norouzi, and David~J. Fleet.
\newblock Video diffusion models, 2022.

\bibitem[Huang et~al.(2024)Huang, He, Yu, Zhang, Si, Jiang, Zhang, Wu, Jin, Chanpaisit, et~al.]{vbench}
Ziqi Huang, Yinan He, Jiashuo Yu, Fan Zhang, Chenyang Si, Yuming Jiang, Yuanhan Zhang, Tianxing Wu, Qingyang Jin, Nattapol Chanpaisit, et~al.
\newblock Vbench: Comprehensive benchmark suite for video generative models.
\newblock In \emph{Proceedings of the IEEE/CVF Conference on Computer Vision and Pattern Recognition}, pages 21807--21818, 2024.

\bibitem[Issachar et~al.(2025)Issachar, Yariv, Benaim, Adi, Lischinski, and Fattal]{issachar2025dype}
Noam Issachar, Guy Yariv, Sagie Benaim, Yossi Adi, Dani Lischinski, and Raanan Fattal.
\newblock Dype: Dynamic position extrapolation for ultra high resolution diffusion.
\newblock \emph{arXiv preprint arXiv:2510.20766}, 2025.

\bibitem[Kong et~al.(2024)Kong, Tian, Zhang, Min, Dai, Zhou, Xiong, Li, Wu, Zhang, et~al.]{kong2024hunyuanvideo}
Weijie Kong, Qi Tian, Zijian Zhang, Rox Min, Zuozhuo Dai, Jin Zhou, Jiangfeng Xiong, Xin Li, Bo Wu, Jianwei Zhang, et~al.
\newblock Hunyuanvideo: A systematic framework for large video generative models.
\newblock \emph{arXiv preprint arXiv:2412.03603}, 2024.

\bibitem[Kye et~al.(2025)Kye, Roh, Ko, Eom, and Oh]{ace}
Dahyeon Kye, Changhyun Roh, Sukhun Ko, Chanho Eom, and Jihyong Oh.
\newblock Acevfi: A comprehensive survey of advances in video frame interpolation, 2025.

\bibitem[Liu et~al.(2024)Liu, Wang, Cao, Jia, and Huang]{understandingediting}
Bingyan Liu, Chengyu Wang, Tingfeng Cao, Kui Jia, and Jun Huang.
\newblock Towards understanding cross and self-attention in stable diffusion for text-guided image editing, 2024.

\bibitem[Liu et~al.(2023{\natexlab{a}})Liu, Zhang, Li, Lin, and Jia]{videop2p}
Shaoteng Liu, Yuechen Zhang, Wenbo Li, Zhe Lin, and Jiaya Jia.
\newblock Video-p2p: Video editing with cross-attention control, 2023{\natexlab{a}}.

\bibitem[Liu et~al.(2023{\natexlab{b}})Liu, Yan, Zhang, An, Qiu, and Lin]{scalinglaw}
Xiaoran Liu, Hang Yan, Shuo Zhang, Chenxin An, Xipeng Qiu, and Dahua Lin.
\newblock Scaling laws of rope-based extrapolation.
\newblock \emph{arXiv preprint arXiv:2310.05209}, 2023{\natexlab{b}}.

\bibitem[Ma et~al.(2022)Ma, Xu, Sun, Yan, Zhang, and Ji]{xclip}
Yiwei Ma, Guohai Xu, Xiaoshuai Sun, Ming Yan, Ji Zhang, and Rongrong Ji.
\newblock X-clip: End-to-end multi-grained contrastive learning for video-text retrieval, 2022.

\bibitem[Park et~al.(2025)Park, Choi, Jun, and Hwang]{wave}
Jiwoo Park, Tae~Eun Choi, Youngjun Jun, and Seong~Jae Hwang.
\newblock Wave: Warp-based view guidance for consistent novel view synthesis using a single image, 2025.

\bibitem[Parmar et~al.(2023)Parmar, Singh, Zhang, Li, Lu, and Zhu]{pix2pixzero}
Gaurav Parmar, Krishna~Kumar Singh, Richard Zhang, Yijun Li, Jingwan Lu, and Jun-Yan Zhu.
\newblock Zero-shot image-to-image translation, 2023.

\bibitem[Peng et~al.(2023)Peng, Quesnelle, Fan, and Shippole]{peng2023yarn}
Bowen Peng, Jeffrey Quesnelle, Honglu Fan, and Enrico Shippole.
\newblock Yarn: Efficient context window extension of large language models.
\newblock \emph{arXiv preprint arXiv:2309.00071}, 2023.

\bibitem[Peng et~al.(2025)Peng, Zheng, Shen, Young, Guo, Wang, Xu, Liu, Jiang, Li, Wang, Ye, Ren, Ma, Liang, Lian, Wu, Zhong, Li, Gong, Lei, Cheng, Zhang, Li, Zhang, Hu, Huang, Wang, Zhao, Wang, Wei, and You]{opensora}
Xiangyu Peng, Zangwei Zheng, Chenhui Shen, Tom Young, Xinying Guo, Binluo Wang, Hang Xu, Hongxin Liu, Mingyan Jiang, Wenjun Li, Yuhui Wang, Anbang Ye, Gang Ren, Qianran Ma, Wanying Liang, Xiang Lian, Xiwen Wu, Yuting Zhong, Zhuangyan Li, Chaoyu Gong, Guojun Lei, Leijun Cheng, Limin Zhang, Minghao Li, Ruijie Zhang, Silan Hu, Shijie Huang, Xiaokang Wang, Yuanheng Zhao, Yuqi Wang, Ziang Wei, and Yang You.
\newblock Open-sora 2.0: Training a commercial-level video generation model in \$200k, 2025.

\bibitem[Pont-Tuset et~al.(2018)Pont-Tuset, Perazzi, Caelles, Arbeláez, Sorkine-Hornung, and Gool]{davis}
Jordi Pont-Tuset, Federico Perazzi, Sergi Caelles, Pablo Arbeláez, Alex Sorkine-Hornung, and Luc~Van Gool.
\newblock The 2017 davis challenge on video object segmentation, 2018.

\bibitem[Radford et~al.(2021)Radford, Kim, Hallacy, Ramesh, Goh, Agarwal, Sastry, Askell, Mishkin, Clark, et~al.]{clip}
Alec Radford, Jong~Wook Kim, Chris Hallacy, Aditya Ramesh, Gabriel Goh, Sandhini Agarwal, Girish Sastry, Amanda Askell, Pamela Mishkin, Jack Clark, et~al.
\newblock Learning transferable visual models from natural language supervision.
\newblock In \emph{International conference on machine learning}, pages 8748--8763. PmLR, 2021.

\bibitem[Reda et~al.(2022)Reda, Kontkanen, Tabellion, Sun, Pantofaru, and Curless]{reda2022film}
Fitsum Reda, Janne Kontkanen, Eric Tabellion, Deqing Sun, Caroline Pantofaru, and Brian Curless.
\newblock Film: Frame interpolation for large motion.
\newblock In \emph{European Conference on Computer Vision}, pages 250--266. Springer, 2022.

\bibitem[Rombach et~al.(2022)Rombach, Blattmann, Lorenz, Esser, and Ommer]{ldm}
Robin Rombach, Andreas Blattmann, Dominik Lorenz, Patrick Esser, and Björn Ommer.
\newblock High-resolution image synthesis with latent diffusion models, 2022.

\bibitem[Shrout and Fleiss(1979)]{icc}
Patrick~E Shrout and Joseph~L Fleiss.
\newblock Intraclass correlations: uses in assessing rater reliability.
\newblock \emph{Psychological bulletin}, 86\penalty0 (2):\penalty0 420, 1979.

\bibitem[Stergiou(2024)]{lavib}
Alexandros Stergiou.
\newblock Lavib: A large-scale video interpolation benchmark, 2024.

\bibitem[Su et~al.(2023)Su, Lu, Pan, Murtadha, Wen, and Liu]{rope}
Jianlin Su, Yu Lu, Shengfeng Pan, Ahmed Murtadha, Bo Wen, and Yunfeng Liu.
\newblock Roformer: Enhanced transformer with rotary position embedding, 2023.

\bibitem[Unterthiner et~al.(2018)Unterthiner, Van~Steenkiste, Kurach, Marinier, Michalski, and Gelly]{fvd}
Thomas Unterthiner, Sjoerd Van~Steenkiste, Karol Kurach, Raphael Marinier, Marcin Michalski, and Sylvain Gelly.
\newblock Towards accurate generative models of video: A new metric \& challenges.
\newblock \emph{arXiv preprint arXiv:1812.01717}, 2018.

\bibitem[Vaswani et~al.(2017)Vaswani, Shazeer, Parmar, Uszkoreit, Jones, Gomez, Kaiser, and Polosukhin]{attention}
Ashish Vaswani, Noam Shazeer, Niki Parmar, Jakob Uszkoreit, Llion Jones, Aidan~N Gomez, {\L}ukasz Kaiser, and Illia Polosukhin.
\newblock Attention is all you need.
\newblock \emph{Advances in neural information processing systems}, 30, 2017.

\bibitem[Wang et~al.(2025)Wang, Zhou, Curless, Kemelmacher-Shlizerman, Holynski, and Seitz]{gi}
Xiaojuan Wang, Boyang Zhou, Brian Curless, Ira Kemelmacher-Shlizerman, Aleksander Holynski, and Steven~M. Seitz.
\newblock Generative inbetweening: Adapting image-to-video models for keyframe interpolation, 2025.

\bibitem[Wang et~al.(2004)Wang, Bovik, Sheikh, and Simoncelli]{ssim}
Zhou Wang, Alan~C Bovik, Hamid~R Sheikh, and Eero~P Simoncelli.
\newblock Image quality assessment: from error visibility to structural similarity.
\newblock \emph{IEEE transactions on image processing}, 13\penalty0 (4):\penalty0 600--612, 2004.

\bibitem[Wei et~al.(2025)Wei, Liu, Zang, Dong, Zhang, Cao, Tong, Duan, Guo, Wang, et~al.]{wei2025videorope}
Xilin Wei, Xiaoran Liu, Yuhang Zang, Xiaoyi Dong, Pan Zhang, Yuhang Cao, Jian Tong, Haodong Duan, Qipeng Guo, Jiaqi Wang, et~al.
\newblock Videorope: What makes for good video rotary position embedding?
\newblock \emph{arXiv preprint arXiv:2502.05173}, 2025.

\bibitem[Wu et~al.(2025)Wu, Siarohin, Menapace, Skorokhodov, Fang, Chordia, Gilitschenski, and Tulyakov]{wu2025mind}
Ziyi Wu, Aliaksandr Siarohin, Willi Menapace, Ivan Skorokhodov, Yuwei Fang, Varnith Chordia, Igor Gilitschenski, and Sergey Tulyakov.
\newblock Mind the time: Temporally-controlled multi-event video generation.
\newblock In \emph{Proceedings of the Computer Vision and Pattern Recognition Conference}, pages 23989--24000, 2025.

\bibitem[Yang et~al.(2024{\natexlab{a}})]{qwen}
An Yang et~al.
\newblock Qwen2 technical report, 2024{\natexlab{a}}.

\bibitem[Yang et~al.(2025)Yang, Kwon, and Ye]{vibid}
Serin Yang, Taesung Kwon, and Jong~Chul Ye.
\newblock Vibidsampler: Enhancing video interpolation using bidirectional diffusion sampler, 2025.

\bibitem[Yang et~al.(2024{\natexlab{b}})Yang, Teng, Zheng, Ding, Huang, Xu, Yang, Hong, Zhang, Feng, et~al.]{cogvideox}
Zhuoyi Yang, Jiayan Teng, Wendi Zheng, Ming Ding, Shiyu Huang, Jiazheng Xu, Yuanming Yang, Wenyi Hong, Xiaohan Zhang, Guanyu Feng, et~al.
\newblock Cogvideox: Text-to-video diffusion models with an expert transformer.
\newblock \emph{arXiv preprint arXiv:2408.06072}, 2024{\natexlab{b}}.

\bibitem[Yuan et~al.(2025)Yuan, Wang, Sun, Zhang, and Lin]{tarsier}
Liping Yuan, Jiawei Wang, Haomiao Sun, Yuchen Zhang, and Yuan Lin.
\newblock Tarsier2: Advancing large vision-language models from detailed video description to comprehensive video understanding, 2025.

\bibitem[Zhang et~al.(2024)Zhang, Dong, Zang, Cao, Qian, Chen, Guo, Duan, Wang, Ouyang, Zhang, Zhang, Li, Gao, Sun, Zhang, Li, Li, Wang, Yan, He, Zhang, Chen, Dai, Qiao, Lin, and Wang]{internlm}
Pan Zhang, Xiaoyi Dong, Yuhang Zang, Yuhang Cao, Rui Qian, Lin Chen, Qipeng Guo, Haodong Duan, Bin Wang, Linke Ouyang, Songyang Zhang, Wenwei Zhang, Yining Li, Yang Gao, Peng Sun, Xinyue Zhang, Wei Li, Jingwen Li, Wenhai Wang, Hang Yan, Conghui He, Xingcheng Zhang, Kai Chen, Jifeng Dai, Yu Qiao, Dahua Lin, and Jiaqi Wang.
\newblock Internlm-xcomposer-2.5: A versatile large vision language model supporting long-contextual input and output, 2024.

\bibitem[Zhang et~al.(2018)Zhang, Isola, Efros, Shechtman, and Wang]{lpips}
Richard Zhang, Phillip Isola, Alexei~A Efros, Eli Shechtman, and Oliver Wang.
\newblock The unreasonable effectiveness of deep features as a perceptual metric.
\newblock In \emph{Proceedings of the IEEE conference on computer vision and pattern recognition}, pages 586--595, 2018.

\bibitem[Zhang et~al.(2025)Zhang, Wu, Li, Li, Ma, Liu, and Li]{llavavideo}
Yuanhan Zhang, Jinming Wu, Wei Li, Bo Li, Zejun Ma, Ziwei Liu, and Chunyuan Li.
\newblock Llava-video: Video instruction tuning with synthetic data, 2025.

\bibitem[Zhou et~al.(2024)Zhou, Zhou, Cheng, Feng, and Hou]{storydiff}
Yupeng Zhou, Daquan Zhou, Ming-Ming Cheng, Jiashi Feng, and Qibin Hou.
\newblock Storydiffusion: Consistent self-attention for long-range image and video generation, 2024.

\bibitem[Zhu et~al.(2024)Zhu, Ren, Wang, Wu, and Zuo]{fcvg}
Tianyi Zhu, Dongwei Ren, Qilong Wang, Xiaohe Wu, and Wangmeng Zuo.
\newblock Generative inbetweening through frame-wise conditions-driven video generation, 2024.

\end{thebibliography}
}
\clearpage
\onecolumn
\appendix

\setcounter{section}{0}
\setcounter{figure}{0}
\setcounter{table}{0}
\setcounter{equation}{0}

\renewcommand{\thesection}{S\arabic{section}}
\renewcommand{\thefigure}{S\arabic{figure}}
\renewcommand{\thetable}{S\arabic{table}}
\renewcommand{\theequation}{S\arabic{equation}}

\begin{center}
    {\LARGE \textbf{Supplementary Material}}\\[0.5em]
\end{center}
\vspace{1em}


\section{Additional Resources}
The implemented code for our method is presented in the \texttt{code} folder in the supplementary material. 
We also present \textit{all} our result for the 81-frame along with the result videos of the baseline Wan~\cite{wan} in the \texttt{videos} folder in the supplementary material.
In addition, a sample dataset of our TGI-Bench is included in the \texttt{TGI-Bench} folder.

\section{Evaluation Details}
\subsection{Experimental Details}
All experiments were conducted on an NVIDIA RTX A6000 GPU (48GB VRAM) using mixed precision (bfloat16). 
We utilized the Wan2.1-FLF2V-14B-720P checkpoint~\cite{wan}, a 14B-parameter diffusion transformer model, UMT5-XXL text encoder, VAE decoder, and XLM-RoBERTa-Large vision encoder which can all be accessed through Wan2.1\footnote{https://github.com/Wan-Video/Wan2.1}.
Inference was performed using the DiffSynth-Studio\footnote{https://github.com/modelscope/DiffSynth-Studio} framework, which provides efficient pipeline management and automatic VRAM optimization.
In addition, videos were generated at 480×864 resolution with 15 FPS using tiled processing. 
For Stable Video Diffusion–based models, we used the following checkpoints in our experiments: stable-video-diffusion-img2vid-xt for GI~\cite{gi}, ViBiD~\cite{vibid}, TRF~\cite{trf} and stable-video-diffusion-img2vid-xt-1-1 for FCVG~\cite{fcvg}.

\subsection{Video Question Answering Evaluation Details}
To obtain a stable VQA-based alignment score between a generated video and its textual prompt, we evaluate each video using six vision–language models with diverse architectures and visual encoders: qwen2.5-vl-7b~\cite{qwen}, llava-onevision-qwen2-7b-sillavaonevision, internlmxcomposer25-7b~\cite{internlm}, tarsier-recap-7b~\cite{tarsier}, llava-video-7b~\cite{llavavideo}, and gpt-4.1~\cite{gpt4}. 
For each model, we sample video frames using either an FPS-based strategy (Qwen models) or a fixed frame-count strategy (LLaVA, InternLM-XComposer, Tarsier), encode the frames through the model's vision encoder, and compute a binary VQA response to the question \texttt{Does this video show \{caption\}?}. 
Each model produces a probability score for the \texttt{Yes} response, normalized to {\small$[0,1]$} from the logits of the \texttt{Yes} and \texttt{No} tokens (or log-probabilities in the case of gpt-4.1). 
Because individual models exhibit significant variance due to differences in frame sampling, vision encoders, and temporal reasoning ability, we average the scores across all six models to obtain a more reliable and model-agnostic VQA metric.

\subsection{User Study Details}
\begin{figure}[]
  \centering
  \includegraphics[width=0.98\linewidth]{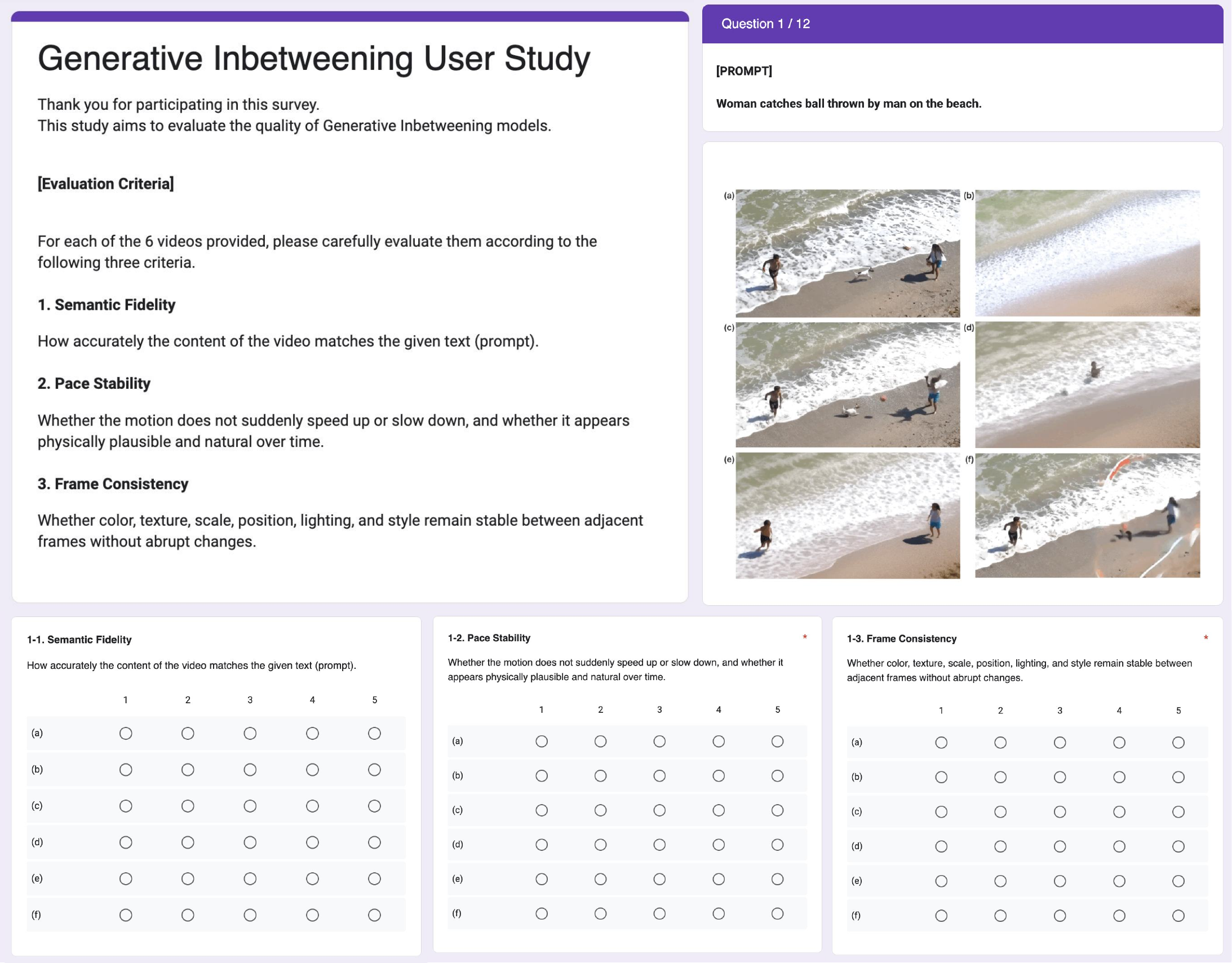}
  \caption{
    \textbf{User study interface used to evaluate our generative inbetweening results.}
    For each text prompt (top right), six candidate videos (a–f) are displayed.
    Participants first read the evaluation criteria (left) and then rate each
    video on a 5-point Likert scale for three dimensions: Semantic Fidelity,
    Pace Stability, and Frame Consistency.
  }
  \vspace{-8pt}
  \label{fig:user_study_interface}
\end{figure}

Figure~\ref{fig:user_study_interface} shows the interface used in our user study, which was conducted with more than 20 participants. 
For each of the 12 questions (about 10\% of TGI-Bench), participants were given a text prompt and 6 video clips generated by each baseline models, whose positions were randomly shuffled to ensure fairness. 
They then rated every clip on semantic fidelity, pace stability, and frame consistency using a five-point Likert scale.
To assess inter-rater reliability, we computed ICC(2,k)~\cite{icc} across the 21 participants, obtaining 0.953 for semantic fidelity, 0.967 for pace stability, and 0.964 for frame consistency.
\section{Additional Experimental Results}

\subsection{Quantitative Results}
In ~\cref{tab:supp_quan1}, we present quantitative results for the 25- and 33-frame sequences.
All settings, except for the number of frames, are identical to those used for the 65- and 81-frame sequences in the main paper.

\setlength{\tabcolsep}{2pt} 
\renewcommand{\arraystretch}{1} 
\setlength{\aboverulesep}{0.5pt}
\setlength{\belowrulesep}{0.5pt}
\setlength{\abovetopsep}{0pt}
\setlength{\belowbottomsep}{0pt}
\setlength{\cmidrulekern}{0pt} 

\begin{table}[h]
\caption{\textbf{Additional Video Generation Evaluation Results.} Quantitative comparison of the baselines and our method on 25, 31 frames. We evaluate video generation quality and fidelity. The best results are in \textbf{bold}, and the second best are \underline{underlined}.} 
\centering
\small
\setlength{\tabcolsep}{3pt}
\begin{tabular*}{\columnwidth}{@{\extracolsep{\fill}} @{} l *{12}{c} @{}}
\toprule
\multirow{2}{*}{Method} &
\multicolumn{6}{c}{\textbf{25-frame}} &
\multicolumn{6}{c}{\textbf{33-frame}} \\
\cmidrule(l{0.2em}r{0.2em}){2-7}\cmidrule(l{0.2em}r{0.2em}){8-13}
& PSNR$\uparrow$ & SSIM$\uparrow$ & LPIPS$\downarrow$ & FID$\downarrow$ & FVD$\downarrow$
& VBench$\uparrow$
& PSNR$\uparrow$ & SSIM$\uparrow$
& LPIPS$\downarrow$ & FID$\downarrow$ & FVD$\downarrow$
& VBench$\uparrow$ \\
\midrule
TRF~\cite{trf}            & 16.734 & 0.5546 & 0.4612 & 104.393 & 0.2749 & 9.473 
                          & 16.603 & 0.5584 & 0.4777 & 118.459 & 0.2893 & 9.147 \\
ViBiDSampler~\cite{vibid} & 17.029 & 0.5686 & 0.4257 & 93.172  & 0.2776 & 9.587 
                          & 16.607 & 0.5574 & 0.4561 & 107.697 & 0.3121 & 9.245 \\
GI~\cite{gi}              & 17.418 & 0.5801 & 0.3972 & 91.884  & 0.2571 & 9.935 
                          & 16.499 & 0.5587 & 0.4470 & 127.957 & 0.2955 & 9.339 \\
FCVG~\cite{fcvg}          & 18.264 & 0.5631 & 0.3859 & 80.276  & 0.2016 & 9.865 
                          & 17.682 & 0.5523 & 0.4083 & 88.814  & 0.2508 & 9.781 \\
Wan~\cite{wan}            & \underline{19.076} & \underline{0.6180} & \underline{0.3430}           
                          & \underline{68.890}  & \underline{0.1821} & \textbf{10.103} 
                          & \underline{18.174} & \underline{0.5953} & \underline{0.3771} 
                          & \underline{74.383} & \underline{0.2409} & \underline{9.915} \\
\hline
\textbf{Ours}             & \textbf{19.557} & \textbf{0.6322} & \textbf{0.3418} 
                          & \textbf{67.888}  & \textbf{0.1682} & \underline{9.991} 
                          & \textbf{18.757} & \textbf{0.6127} & \textbf{0.3669} 
                          & \textbf{70.399}  & \textbf{0.2086} & \textbf{9.918} \\
\bottomrule
\end{tabular*}
\label{tab:supp_quan1}
\end{table}

\subsection{Qualitative Results}
\begin{figure*}[t]
  \centering
    \includegraphics[width=\linewidth]{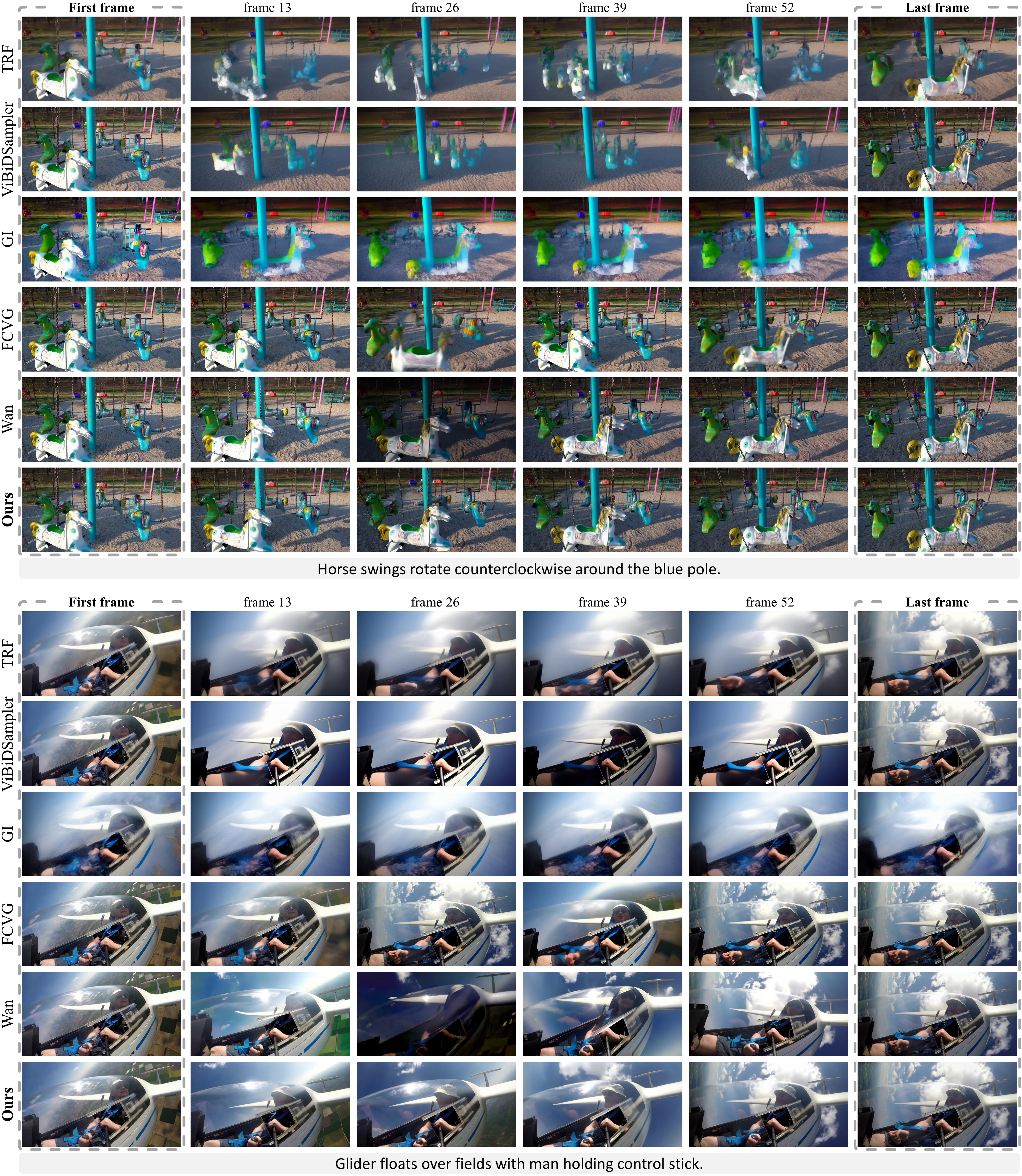}
    \caption{\textbf{Qualitative Results.} 
    In both examples, our method generates consistent frames compared to Wan which shows artifacts or suddenly dimmed scenes. The first four models fail to maintain the object shape for the intermediate frames. 
    }
    \label{fig:suppl_qualitative5}
\end{figure*}

\begin{figure*}[t]
  \centering
    \includegraphics[width=\linewidth]{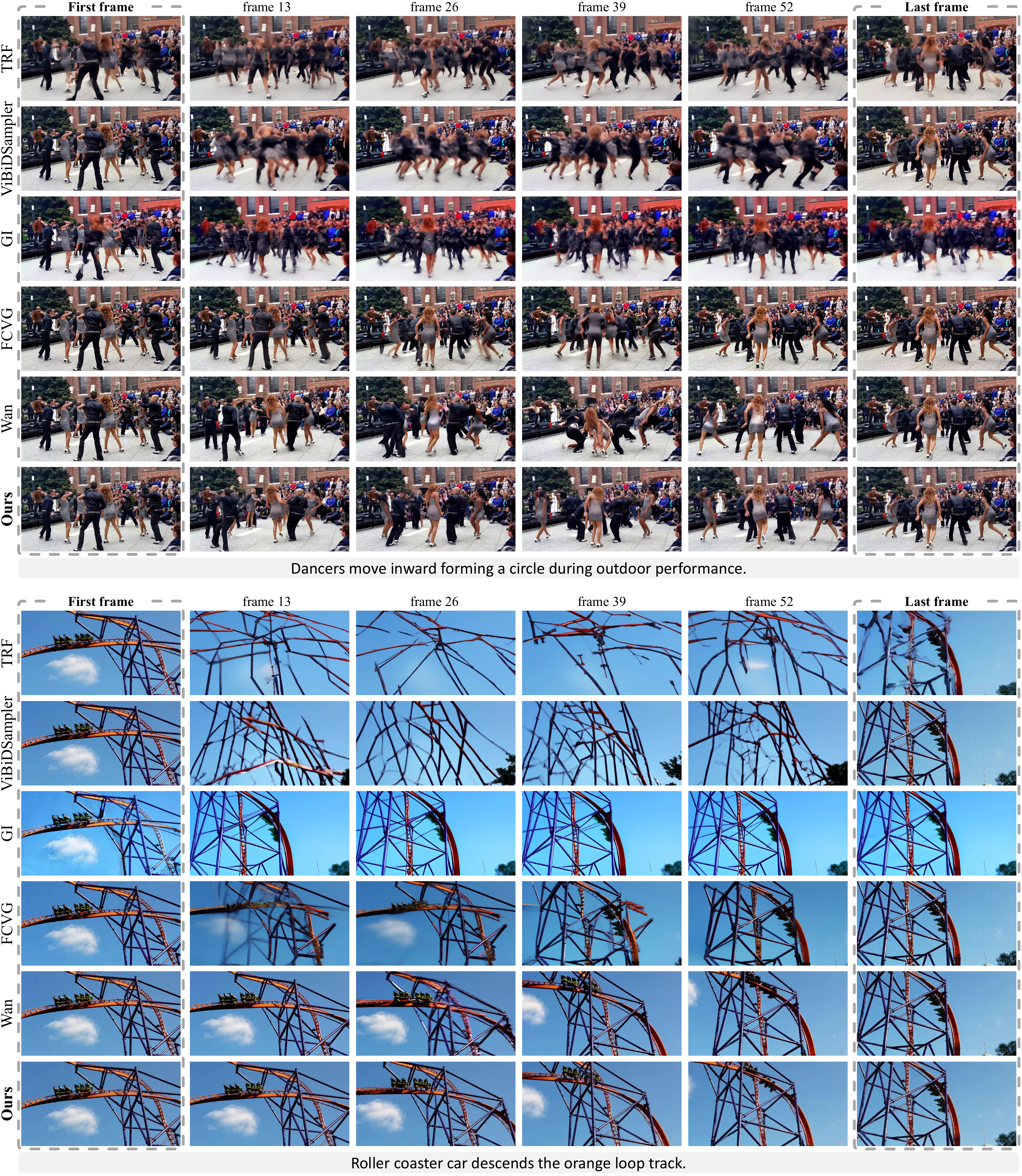}
    \caption{\textbf{Qualitative Results.}
    (Top) Other models, unlike ours, either show blurred objects with inconsistent frames or unnatural motion like Wan in frame 39. Our method shows high semantic fidelity as well as frame consistency through all frames. 
    (Bottom) For the first four models, the structure of the rollercoaster collapses, failing to maintain the shape and style of the keyframes. Our model shows pace stability while maintaining the frame consistency.} 
    \label{fig:suppl_qualitative6}
\end{figure*}

\begin{figure*}[t]
  \centering
    \includegraphics[width=\linewidth]{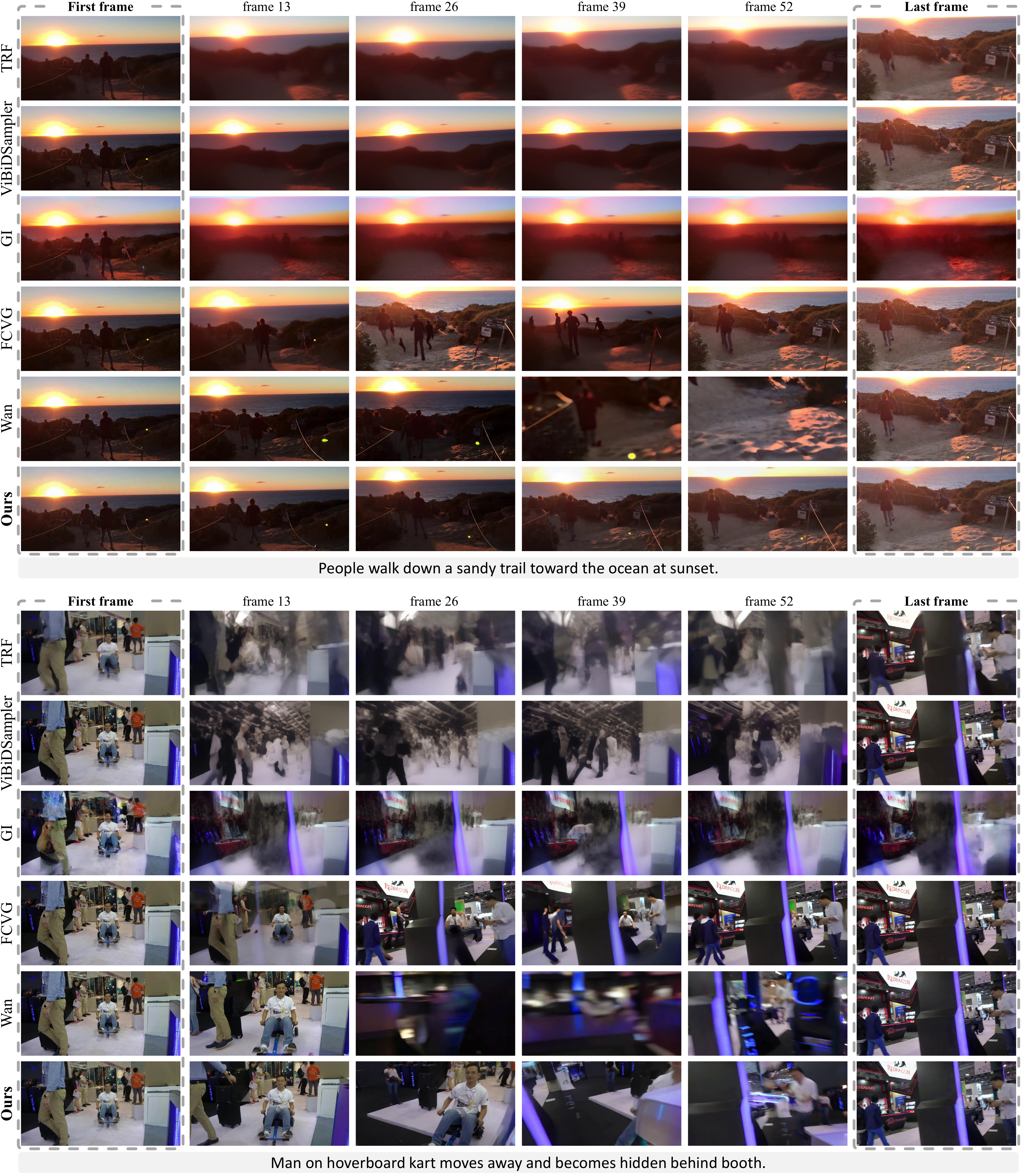}
    \caption{\textbf{Qualitative Results.} Examples showing that our method performs well in highly complex scenes with multiple people and objects, preserving fine details and producing fewer blurred scenes than baseline methods.} 
    \label{fig:suppl_qualitative7}
\end{figure*}

\begin{figure*}[t]
  \centering
    \includegraphics[width=\linewidth]{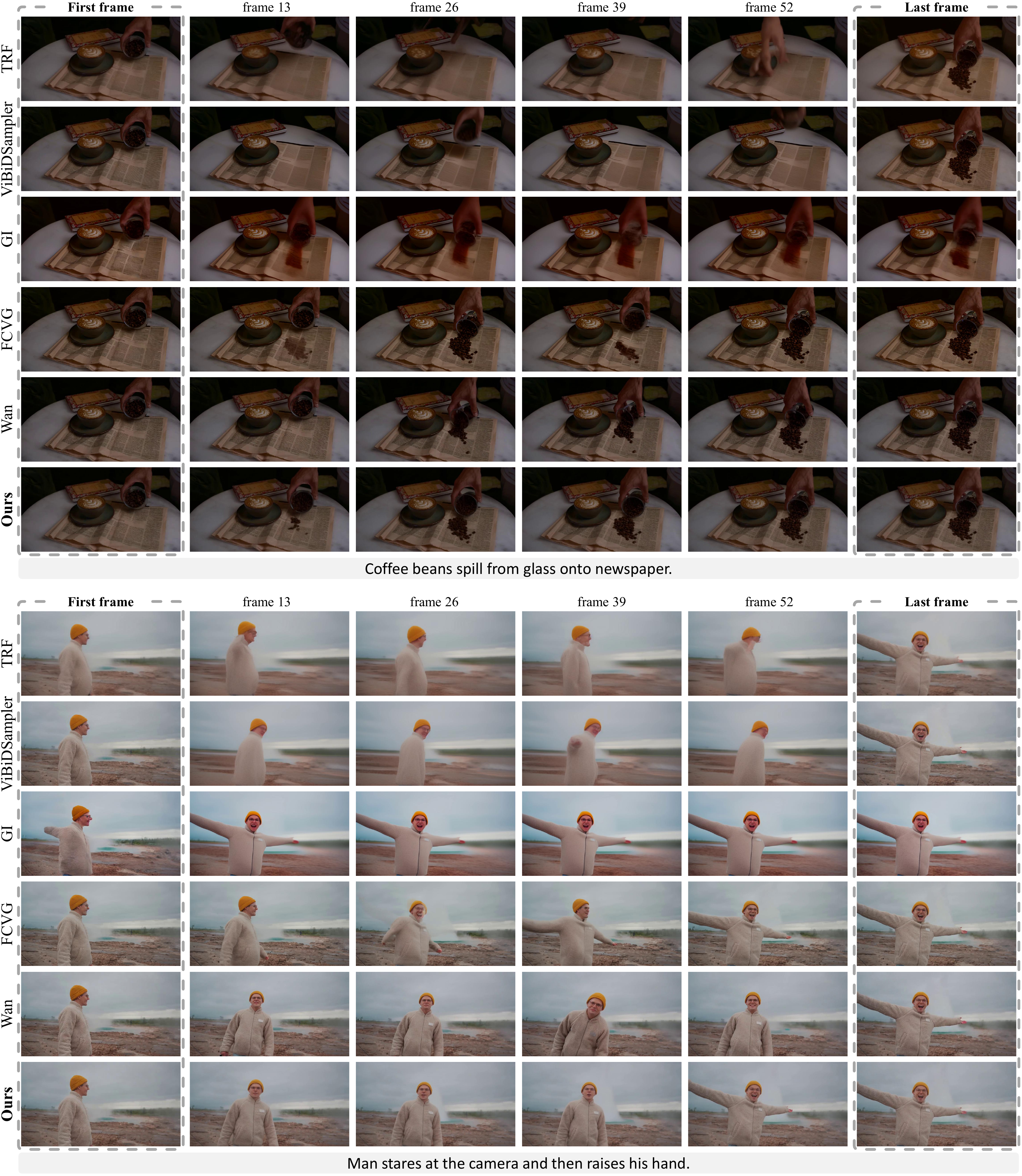}
    \caption{\textbf{Qualitative Results.} 
    (Top) The first four baselines show unstable coffee-spilling pace and temporal inconsistency, while even compared to Wan our method generates more stably paced and temporally coherent motion.
    (Bottom) The first four baselines suffer from blur that distorts the human shape. Compared to Wan, our method maintains a more stable pace and generates more natural motions. }
    \label{fig:suppl_qualitative8}
\end{figure*}

\begin{figure*}[t]
  \centering
    \includegraphics[width=\linewidth]{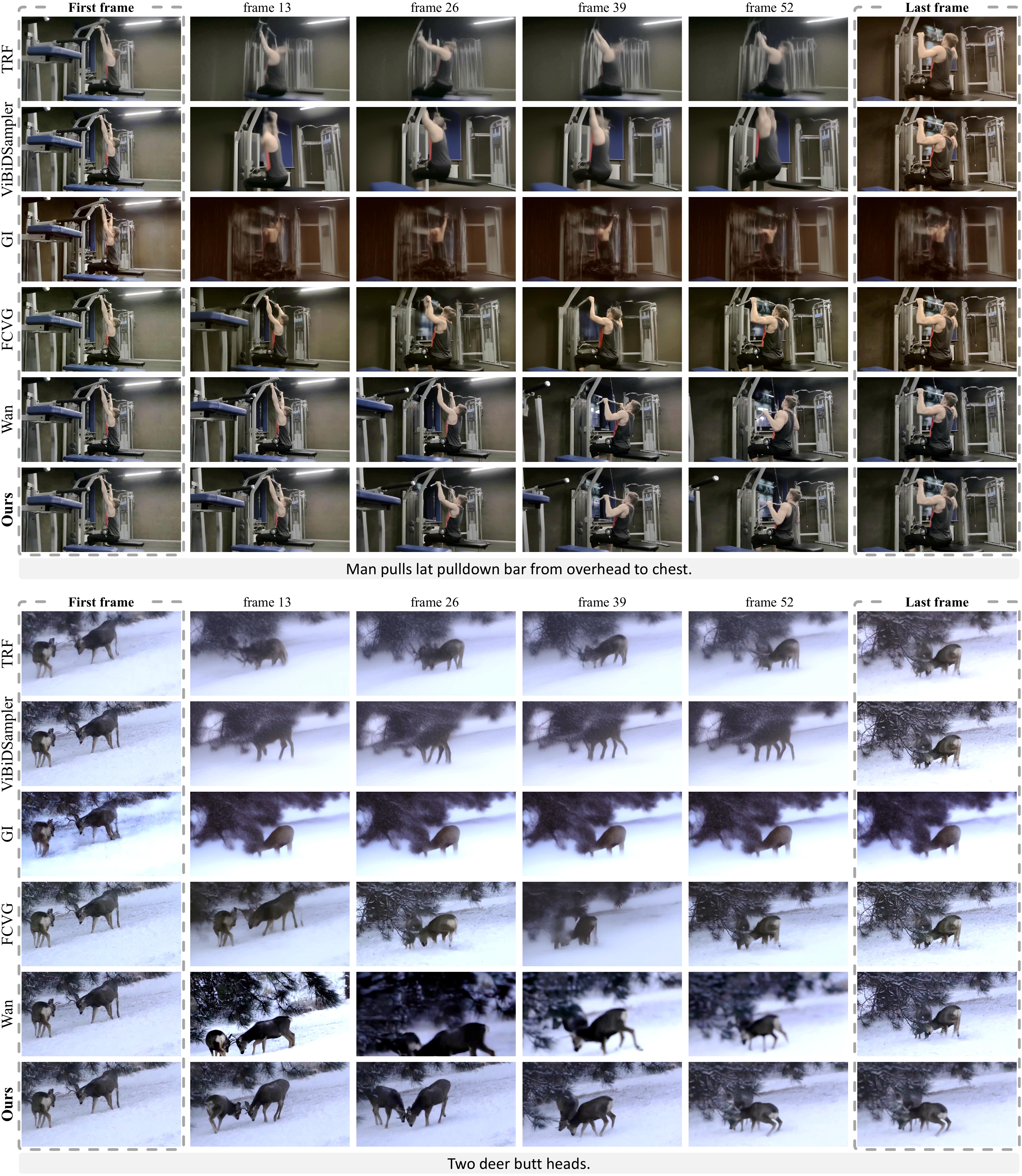}
    \caption{\textbf{Qualitative Results.} 
    (Top) The first four baselines produce overly blurred frames where the human shape is not preserved and even compared to Wan, our method exhibits a more stable motion pace for the man performing lat pulldown.
    (Bottom) While other methods contain several blurred and inconsistent frames, our method generates clearer and more temporally consistent videos.
    For visualization purposes, we uniformly increased the brightness of both examples by 40\%, while leaving all other properties unchanged.
    } 
    \label{fig:suppl_qualitative9}
\end{figure*}

\begin{figure*}[t]
  \centering
    \includegraphics[width=\linewidth]{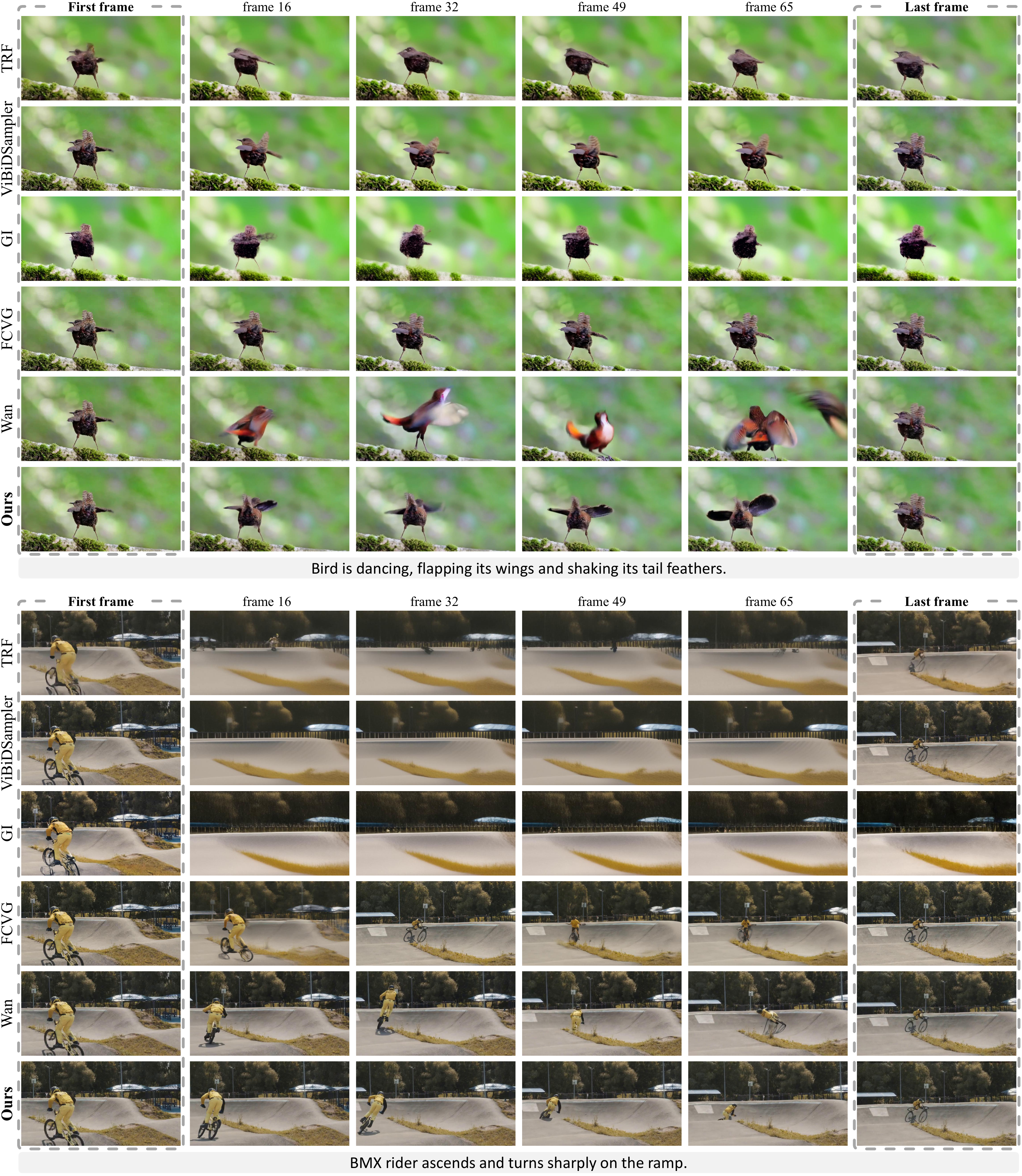}
    \caption{\textbf{Qualitative Results.}
    (Top) For the first four models, there is minimal wing flapping and motion, while our method and Wan show movements. However, Wan fails to maintain frame consistency and semantic fidelity. 
    (Bottom) For Wan, the person on the bicycle goes left in the first few frames but suddenly turns from the right. On the other hand, our method shows consistent pace and consistency in movements. 
    } 
    \label{fig:suppl_qualitative1}
\end{figure*}

\begin{figure*}[t]
  \centering
    \includegraphics[width=\linewidth]{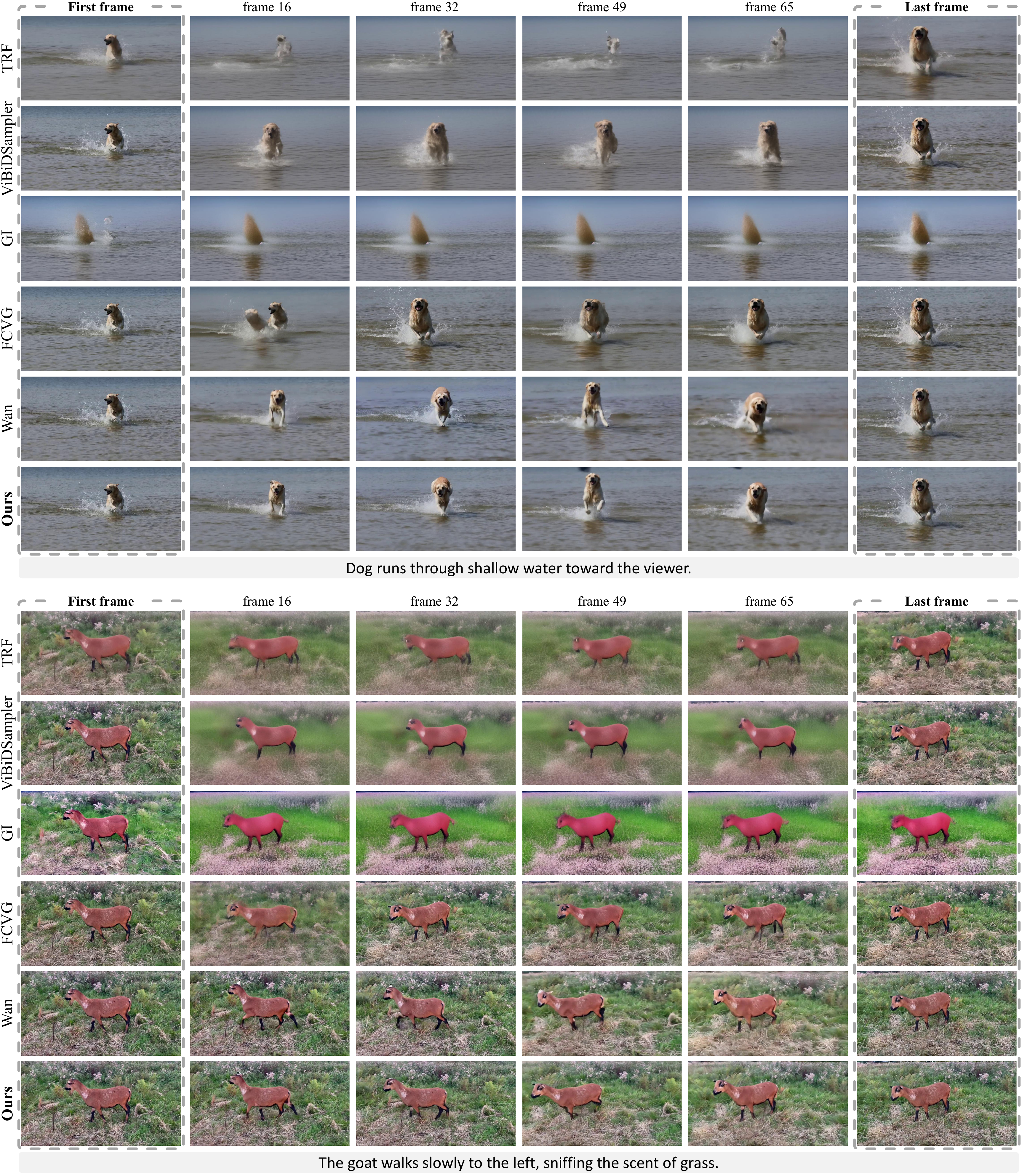}
    \caption{\textbf{Qualitative Results.} The first four models, unlike Wan and ours, fail to maintain the shape of the object as well as the background style through the long frame sequences, showing the importance of correct text prompts in generative inbetweening.} 
    \label{fig:suppl_qualitative2}
\end{figure*}

\begin{figure*}[t]
  \centering
    \includegraphics[width=\linewidth]{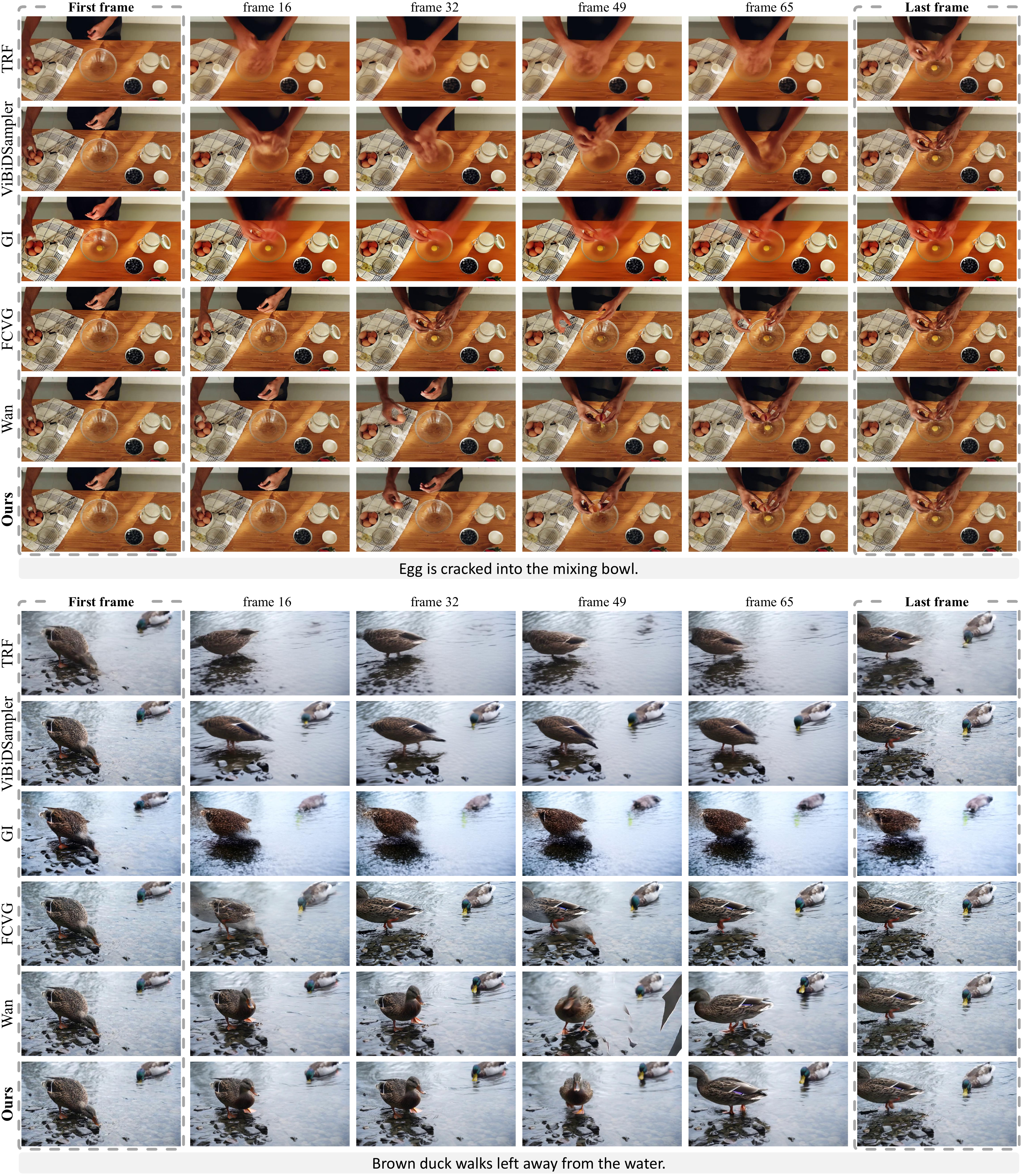}
    \caption{\textbf{Qualitative Results.}
    (Top) The first three models fails to maintain the shape of the hand and egg, while FCVG shows unnatural movement around frame 49 compared to the following two models, Wan and ours. 
    (Bottom) For Wan, an artifact can be observed in frame 49, unlike our method.
    } 
    \label{fig:suppl_qualitative3}
\end{figure*}

\begin{figure*}[t]
  \centering
    \includegraphics[width=\linewidth]{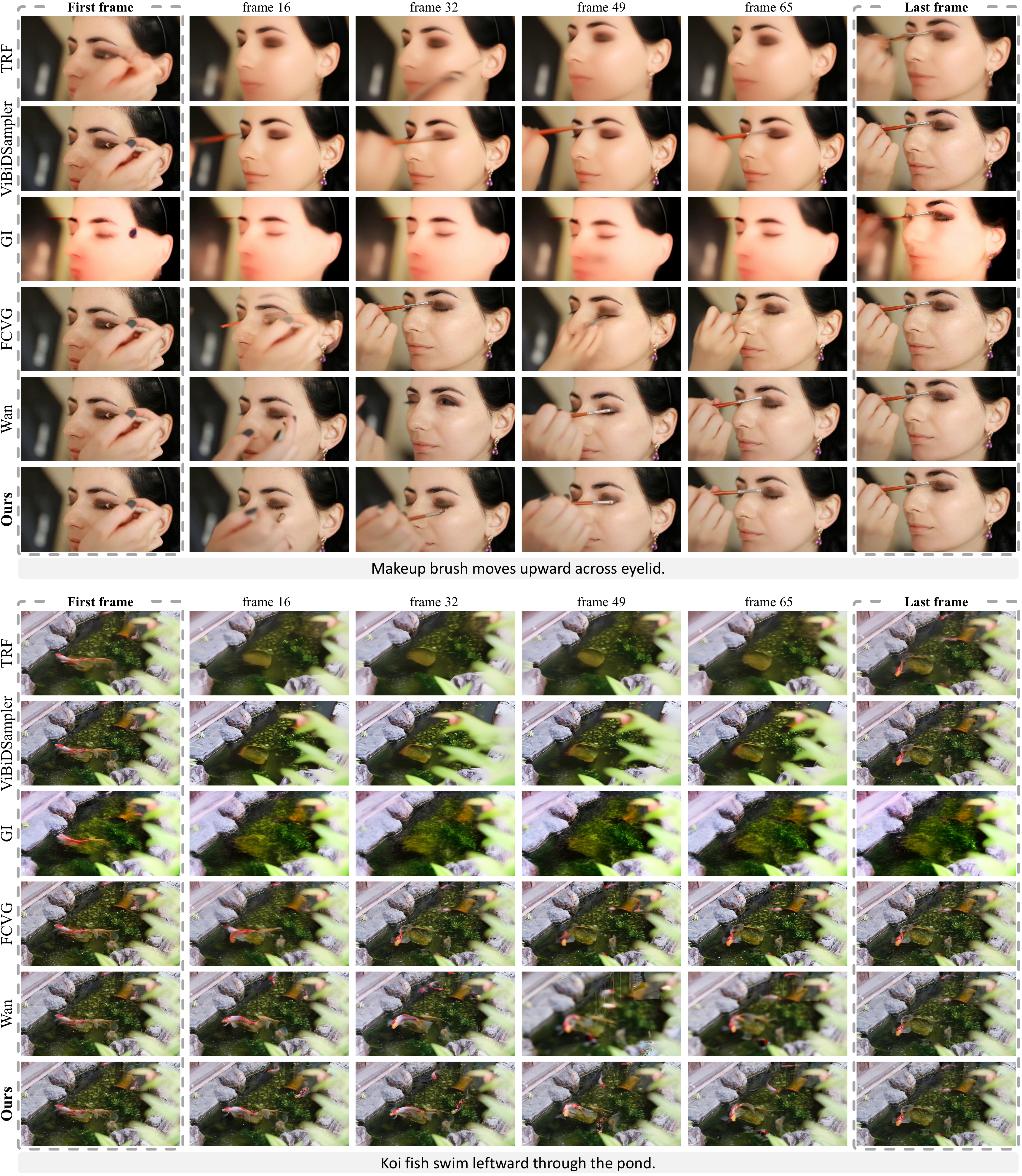}
    \caption{\textbf{Qualitative Results.}
    (Top) While Wan maintains the subject through the long sequence, it does not follow the prompt especially around frame 32. On the other hand, our method faithfully follows the text showing semantic fidelity. 
    (Bottom) Around frame 49-65, Wan shows blurred scene without any context. On the other hand, this problem does not show up on our method.  
    } 
    \label{fig:suppl_qualitative4}
\end{figure*}

We provide additional qualitative results for all our baseline models in Figs.~\ref{fig:suppl_qualitative5}--\ref{fig:suppl_qualitative4}.

\subsection{Hyperparameter Experiment}
We conduct an ablation study on hyperparameters on the two main components of our method, Keyframe-anchored Attention Bias (KAB) and Rescaled Temporal RoPE (ReTRo).
The results are summarized in~\cref{tab:ablation_kab} and~\cref{tab:ablation_retro}. 
Overall, these ablations indicate that our chosen hyperparameters provide a good balance between fidelity and perceptual quality, and that our method is reasonably robust to moderate changes in these values.

For KAB, we experiment over the temporal range $[\beta_t^{\min}, \beta_t^{\max}]$. 
Narrow or overly wide ranges as well as too high or low values generally degrade performance across distortion and perceptual metrics. 
In contrast, our default setting {\small$(0.3 \le \beta_t \le 0.7)$} achieves the best overall scores, yielding clear gains in PSNR, SSIM~\cite{ssim}, and FVD~\cite{fvd} while also improving perceptual quality (VBench~\cite{vbench}). 

To analyze the effect of the ReTRo, we scale the parameters {\small$s_{mid}$} and {\small$s_{edge}$}, which control the relative emphasis on mid-sequence versus boundary frames in the temporal RoPE rescaling. 
Our default configuration {\small$(s_{mid}=0.94,\; s_{edge}=1.06)$} achieves the best performance on most metrics, including PSNR, SSIM, LPIPS~\cite{lpips}, FID~\cite{fid}, and FVD, while maintaining competitive VBench scores. 
The alternative setting {\small$(s_{mid}=0.88,\; s_{edge}=1.06)$} provides the second-best overall performance and slightly higher VBench.

\begin{table}[]
  \centering
  \caption{\textbf{Hyperparameter Experiment Results on KAB.} The best results are in \textbf{bold}, and the second best are \underline{underlined}. \textbf{Ours} denotes the hyperparameters used in our method.}
  \label{tab:ablation_kab}
  \begin{tabular}{lcccccc}
    \toprule
    Hyperparameters & PSNR$\uparrow$ & SSIM$\uparrow$ & LPIPS$\downarrow$ & FID$\downarrow$ & FVD$\downarrow$ & VBench$\uparrow$ \\
    \midrule
    $0.1 \le \beta_t \le 0.5$      & 17.0651 & \underline{0.5859} & 0.3972 & 84.898 & 0.2929 & \textbf{10.237} \\
    $0.5 \le \beta_t \le 0.9$      & 17.100 & 0.5856 & 0.3973 & 85.051 & \underline{0.2839} & \underline{10.230} \\
    $0.5 \le \beta_t \le 0.5$      & \underline{17.107} & 0.5859 & \underline{0.3968} & \underline{83.915} & 0.2865 & 10.203 \\
    $0.1 \le \beta_t \le 0.9$      & 17.072 & 0.5853 & 0.3977 & 84.636 & 0.2887 & 10.208 \\
    \midrule
    \textbf{Ours} $(0.3 \le \beta_t \le 0.7)$ & \textbf{18.169} & \textbf{0.6269} & \textbf{0.3818} & \textbf{77.587} & \textbf{0.2458} & 10.022 \\
    \bottomrule
    \vspace{-15pt}
  \end{tabular}
\end{table}

\begin{table}[]
  \centering 
  \caption{\textbf{Hyperparameter Ablation on ReTRo.} The best results are in \textbf{bold}, and the second best are \underline{underlined}. \textbf{Ours} denotes the hyperparameters used in our method.}
  \label{tab:ablation_retro}
  \makebox[\linewidth][c]{
  \begin{tabular}{l
                  >{\centering\arraybackslash}m{1.0cm}
                  >{\centering\arraybackslash}m{1.0cm}
                  cccccc}
    \toprule
    & \multicolumn{2}{c}{Hyperparameters} & \multicolumn{6}{c}{Metrics} \\
    \cmidrule(lr){2-3} \cmidrule(lr){4-9}
     & $s_{\text{mid}}$ & $s_{\text{edge}}$ &
    PSNR$\uparrow$ & SSIM$\uparrow$ & LPIPS$\downarrow$ &
    FID$\downarrow$ & FVD$\downarrow$ & VBench$\uparrow$ \\
    \midrule
            & 0.94 & 1.12 & 16.964 & 0.5819 & 0.4054 & 90.584 & 0.3005 & \underline{10.157} \\
            & 0.88 & 1.06 & \underline{17.481} & \underline{0.5941} & \underline{0.3842} & \underline{78.739} & \underline{0.2660} & \textbf{10.339} \\
    \midrule
    \textbf{Ours} & 0.94 & 1.06 & \textbf{18.169} & \textbf{0.6269} & \textbf{0.3818} & \textbf{77.587} & \textbf{0.2458} & 10.022 \\
    \bottomrule
    \vspace{-18pt}
  \end{tabular}
  }
\end{table}
\section{Additional Analysis}
\subsection{KAB}
KAB is a method that uses the cross-attention of the keyframes to guide intermediate frames under three conditions: the two keyframes and the text prompt.
Through rigorous experiments in the main paper and in the supplementary material, we have shown that this additional guidance is effective in maintaining both semantic fidelity and pace stability.
However, when the guidance is either too weak or overly strong, it instead degrades these properties, along with the overall video generation quality.

As shown in~\cref{tab:ablation_kab}, our default mid-range setting $(0.3 \le \beta_t \le 0.7)$ clearly outperforms all other ranges on PSNR, SSIM, LPIPS, FID, and FVD. 
Interestingly, ranges biased toward either lower $(0.1 \le \beta_t \le 0.5)$ or higher $(0.5 \le \beta_t \le 0.9)$ scales achieve slightly higher VBench scores, but this comes at the cost of noticeably worse distortion and distributional metrics. 
The very narrow range $(0.5 \le \beta_t \le 0.5)$ yields the second-best FID and LPIPS among the ablated settings, yet still fails to close the gap to our default configuration. 
Taken together, these results suggest that while concentrating guidance at specific diffusion phases can bring marginal gains in certain perceptual aspects, distributing KAB over a moderate mid-range window is crucial for obtaining consistent improvements across both fidelity and perceptual metrics.
Thus, our chosen setting strikes a good balance when applied with a moderate guidance range, which we have empirically demonstrated through our ablation studies.
\begin{figure*}[t]
  \centering
    \includegraphics[width=0.93\linewidth]{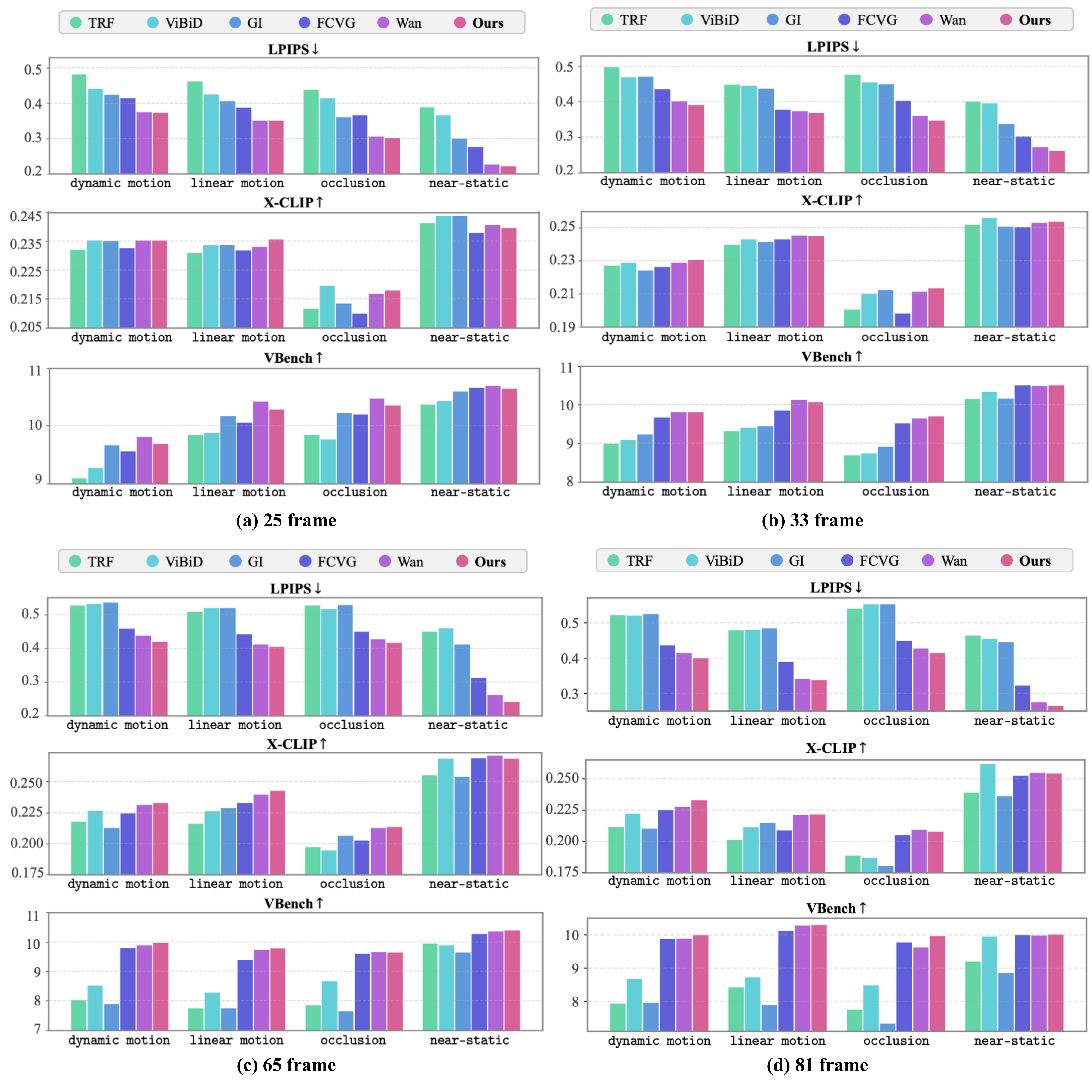}
    \caption{\textbf{Complete Results on Generative Inbetweening Challenge Analysis.} Results for all challenges at frames 25, 33, 65, and 81, including VBench, LPIPS, and X-CLIP scores for each challenge.} 
    \vspace{-10pt}
    \label{fig:challenge_all}
\end{figure*}

\subsection{ReTRo}
ReTRo adaptively modulates RoPE scales along the temporal axis, assigning higher scales to tokens near keyframes to sharpen locality and preserve keyframe content, while using lower scales on intermediate frames to broaden attention and promote temporal consistency. 
In the main paper, we showed that this method is effective in improving both frame consistency and overall video generation quality.

As shown in ~\cref{tab:ablation_retro}, we additionally conducted an ablation study on the ReTRo hyperparameters {\small$s_\text{mid}$} and {\small$s_\text{edge}$}. 
From the results, we found that the parameter setting used in the original paper, {\small$(s_{\text{mid}} = 0.94, s_{\text{edge}} = 1.06)$}, achieved the best performance among the configurations we tested. 
For {\small$s_{\text{edge}}$}, values around 1.10 or higher started to introduce noticeable visual artifacts, while for {\small$s_{\text{mid}}$}, smaller values tended to make the generated videos appear slightly slower in terms of motion. 
Consequently, we adopt {\small$(s_{\text{mid}} = 0.94, s_{\text{edge}} = 1.06)$} as our default setting, as it offers the best trade-off between visual quality and temporal coherence in our experiments. 
However, since these are simple hyperparameters, they can be exposed as user-adjustable parameters, allowing users to dynamically adjust the balance between sharpness, motion speed, and temporal consistency to suit their specific applications.

\subsection{Generative Inbetweening Challenge Analysis}
We additionally present quantitative results for three representative frames (25, 33, and 65) from for further analysis on the generative inbetweening challenges.
The examples are shown in Figs.~\ref{fig:challenge_all}.
These results further confirm that the four challenge categories in TGI-Bench are categorized well in difficulty since most models perform reliably on the \texttt{near-static} cases, whereas performance degrades sharply on the more demanding \texttt{occlusion} and \texttt{dynamic motion} challenges. 
This clear differentiation demonstrates that TGI-Bench is carefully constructed to expose distinct failure modes of GI models, enabling fine-grained diagnosis of model capabilities. 
Consequently, TGI-Bench provides a reliable and informative metric suite for future research, particularly for identifying which generative inbetweening challenges a model handles well and where it struggles.

\section{TGI-Bench Details}
\subsection{Dataset Curation Details}
To construct the TGI-Bench dataset, we prompted GPT-4.1~\cite{gpt4} to generate a text description and a challenge label for each video. The text description often includes information inferred from intermediate frames that are not visible in the provided first and last frames, thereby serving as constraints when a model attempts to generate the missing frames. Inspired by \cite{ace}, the challenge label is categorized into one of four types: \texttt{dynamic motion}, \texttt{linear motion}, \texttt{occlusion}, and \texttt{near-static}, defined as follows:
\begin{itemize}
    \item \textbf{Dynamic motion}: The primary object exhibits nonlinear or complex movement, such as rotation or abrupt directional changes.
    \item \textbf{Linear motion}: The primary object moves in a linear and consistent direction.
    \item \textbf{Occlusion}: The primary object either appears or disappears in the middle of the video due to occlusion.
    \item \textbf{Near-static}: The primary object remains largely stationary with minimal motion.
\end{itemize}

We sourced videos from the DAVIS~\cite{davis} dataset as well as from Pexels and Pixabay\footnote{https://www.pexels.com/, https://www.pixabay.com/}. Videos that were too visually complex to describe succinctly, or that lacked a clearly identifiable primary object, were excluded. For example, we removed videos where geometric patterns changing chaotically or where smoke particles moving randomly without a coherent subject. After this filtering step, we collected a total of 220 videos.
For each video, we selected only the first {\small $F$} frames and discarded videos whose total frame count was less than {\small $F$}. From these, we took frames at indices {\small$\{0, 10, 20, \ldots, \lfloor (F-1)/10 \rfloor, F-1\}$} and provided them to GPT-4.1 along with the prompt in Sec.~\ref{sec:prompt}. The resulting text descriptions and challenge labels were manually reviewed and corrected by the authors to ensure accuracy. 
In particular, GPT’s generic label \texttt{large motion} was refined into the more specific categories of \texttt{dynamic motion} and \texttt{linear motion}. This process was repeated for {\small $F \in \{25, 33, 65, 81\}$}, yielding four distinct subsets of the dataset.

\subsection{GPT Prompt}
\label{sec:prompt}
In this section, we present the detailed prompts provided to GPT-4.1. 
By default, we feed the model the concatenation of \texttt{SYSTEM\_PROMPT} and \texttt{USER\_PROMPT\_BASE}. 
For videos where the model produced outputs that did not follow the intended format, 
we additionally concatenate \texttt{RETRY\_PROMPT} to the input.

\begin{Verbatim}[
    fontsize=\scriptsize,
    breaklines=true,
    breakanywhere=true,
    frame=single,
    framesep=2mm,
    framerule=0.4pt,
    rulecolor=\color{gray}
]

SYSTEM_PROMPT = """
You are a caption generator for a Video Frame Interpolation (VFI) evaluation set.
INPUT: two endpoint images - A (start) and B (end), optional reference images R_i sampled between A and B, and optional reference text (prompts.txt).
TASKS
1) Briefly describe A and B (visible, objective facts; ≤ 20 words each).
2) Classify the challenge as exactly one of:
   - Large motion
   - Occlusion
   - Near-static
   If ambiguous, tiebreaker: Occlusion > Large motion > Near-static.
3) Generate exactly ONE caption that best describes the plausible situation across A->B.
   - Prefer wording and nouns from the reference prompts when correct.
   - If the reference contains mistakes or conflicts with the images, FIX them in your caption.
CAPTION STYLE (strict)
- English only. <=12 words. One simple clause.
- You may include direction if clearly implied by the endpoints.
- No commas/semicolons. Avoid: and, then, while, as, because, so, therefore, hence.
- No meta words: relative, compared, background, foreground, camera, optical flow, frame, endpoint(s).
- No hedging or subjective words.
- Do NOT mention A/B or frames.
CONSISTENCY
- Match direction/size/visibility in endpoints.
- Use "emerges/appears/enters" only if absent at A and present at B.
OUTPUT JSON ONLY:
{
  "first_image_desc": "< ≤20 words >",
  "last_image_desc": "< ≤20 words >",
  "challenge": "Large motion | Occlusion | Near-static",
  "caption": "< ≤12 words >"
}
""".strip()


USER_PROMPT_BASE = """
Images follow in order: A (start), zero or more reference images R_i between A and B, then B (end).
Return JSON ONLY following the schema. English only.
""".strip()

RETRY_PROMPT = """
Return VALID JSON ONLY with keys:
first_image_desc, last_image_desc, challenge, caption.
Choose one: Large motion | Occlusion | Near-static.
One caption only; ≤12 words; one clause; obey all style rules.
""".strip()
\end{Verbatim}

\section{Limitation}

\begin{figure}[h]
  \centering
  \includegraphics[width=\linewidth]{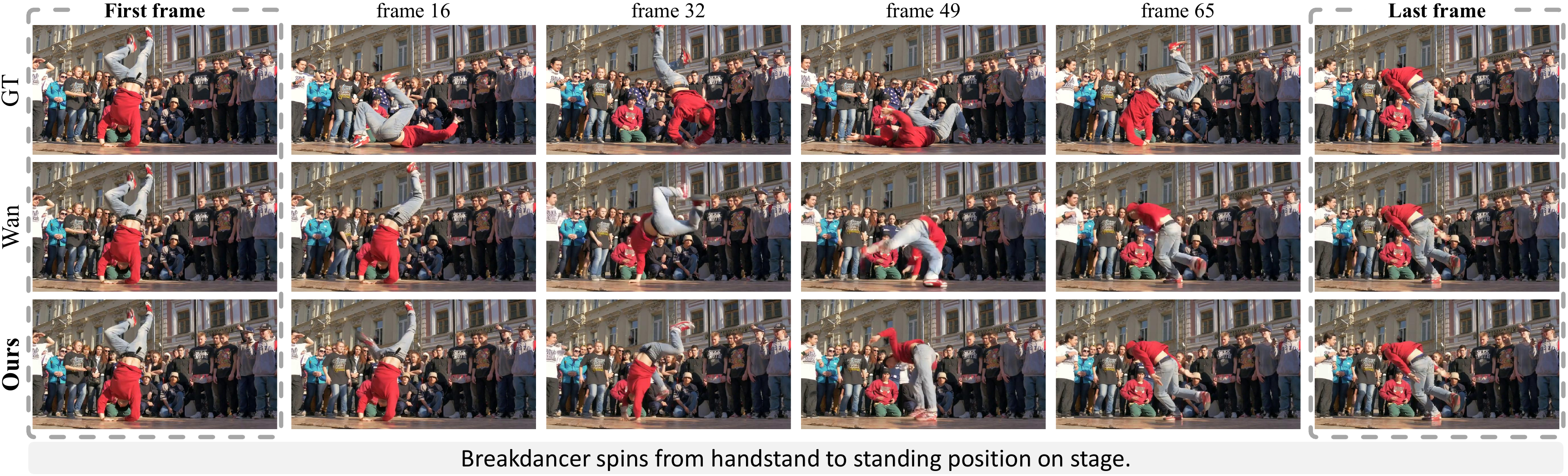}
  \caption{\textbf{Limitation.} Because our training-free plug-in is bounded by the generative capacity of Wan, it can only partially correct the severely distorted motion and geometry seen in the breakdancing example, leaving residual shaky and unnatural motion.}
  \vspace{-8pt}
  \label{fig:limitation}
\end{figure}

Although our method is a simple, training-free plug-in that can be readily applied to existing DiT-based models, it is inherently bounded by the generative capacity of the underlying baseline, Wan~\cite{wan}. 
In challenging cases where the base model already produces severely distorted motion or object geometry over most frames, our approach has limited ability to fully recover a plausible video. 
For example, as shown in~\cref{fig:limitation}, the breakdancing subject exhibits persistent shaky and unnatural motion across time, and these artifacts are only partially mitigated by our method. 
We regard this as a natural limitation of training-free refinement methods and as a promising direction for future work on jointly improving both the base generator and the inbetweening modules.

\section{Ethical Considerations}
TGI-Bench builds on publicly available video dataset Davis~\cite{davis} and open-source video websites Pexels and Pixabay that permit research use. 
We do not collect new data of human subjects, nor do we attempt to infer or annotate sensitive attributes (e.g., identity, race, health, or political views). 
Any videos containing people are used only for generic motion and scene understanding, and are treated as anonymous visual content.

We conducted a small-scale human evaluation with more than 20 participants to compare perceptual quality and consistency, under strict ethical considerations. 
The study followed a double-blind setup, where participants were unaware of the underlying methods being compared, and experimenters did not have access to any identifying information about individual participants. 
All participants volunteered to take part in the study and were informed that the evaluation was conducted solely for academic research purposes. 
No personal information was collected beyond basic platform metadata, and responses were analyzed only in aggregate. 
No offensive, violent, or explicit prompts were used in any of our experiments.

Generative inbetweening can, in principle, be misused for deceptive or non-consensual content (e.g., manipulated videos). 
We explicitly prohibit such uses. 
Our method is presented for research purposes, and any future release of code, models, or benchmarks should be accompanied by clear usage guidelines and restrictions, encouraging applications such as animation, content restoration, and creative tools while discouraging privacy-invasive or harmful deployments.

\clearpage
\end{document}